\documentclass{article} 
\usepackage{iclr2025_conference,times}

\usepackage{amsmath,amsfonts,bm}

\def\eqref#1{equation~\ref{#1}}

\def\1{\bm{1}}

\DeclareMathAlphabet{\mathsfit}{\encodingdefault}{\sfdefault}{m}{sl}
\SetMathAlphabet{\mathsfit}{bold}{\encodingdefault}{\sfdefault}{bx}{n}

\usepackage{hyperref}
\usepackage{url}
\usepackage[pdftex]{graphicx}

\usepackage{placeins} 

\iclrfinalcopy

\usepackage{amsthm}
\usepackage{amsmath}
\newtheorem{theorem}{Theorem}
\numberwithin{theorem}{section} 
\newtheorem{prop}[theorem]{Proposition}

\usepackage{wrapfig}

\title{Fast kernel methods: Sobolev, physics-Informed, and additive models} 

\author{Nathan Doumèche \\
LPSM\\
Sorbonne Universit\'e \\
\And
Francis Bach \\
INRIA Paris\\
\And
Gérard Biau\\
LPSM\\
Sorbonne Universit\'e \\
\AND
Claire Boyer\\
LMO\\
Université Paris-Saclay \\
\texttt{\{nathan.doumeche, gerard.biau\}@sorbonne-universite.fr} \\
\texttt{claire.boyer@universite-paris-saclay.fr} \\
\texttt{francis.bach@inria.fr} 
}

\begin{document}

\maketitle

\begin{abstract}



Kernel methods are powerful tools in statistical learning, but their cubic complexity in the sample size $n$ limits their use on large-scale datasets. In this work, we introduce a scalable framework for kernel regression with $\mathcal{O}(n \log n)$ complexity, fully leveraging GPU acceleration. The approach is based on a Fourier representation of kernels combined with non-uniform fast Fourier transforms (NUFFT), enabling exact, fast, and memory-efficient computations. 
We instantiate our framework in three settings: Sobolev kernel regression, physics-informed regression, and additive models. When known, the proposed estimators are shown to achieve  minimax convergence rates, consistent with classical kernel theory. Empirical results demonstrate that our methods can process up to tens of billions of samples within minutes, providing both statistical accuracy and computational scalability. 
These contributions establish a flexible approach, paving the way for the routine application of kernel methods in large-scale learning tasks.
\end{abstract}


\section{Introduction}

\paragraph{Kernel methods.} Kernel methods play a central role in statistical learning and nonparametric regression, providing flexible and theoretically sound tools for modeling complex data structures \citep{steinwart2008}. Despite their appeal, their practical application is often hindered by severe computational constraints: standard kernel ridge regression has a time complexity of $\mathcal{O}(n^3)$ and a memory cost of $\mathcal{O}(n^2)$, making it unsuitable for large-scale datasets \citep[][Chapter~7]{bach2024learning}. Several approximation schemes, such as Nyström methods and random feature expansions, have been proposed to reduce this burden, reducing the computational cost to $\mathcal{O}(n^{3/2})$. However, these approximations introduce additional error terms, complicating theoretical guarantees and often degrading empirical performance \citep{Meanti2020kernel}.


\paragraph{Contributions.} We show that a broad class of kernel estimators can be trained exactly at a substantially reduced computational cost, achieving $\mathcal{O}(n \log n)$ complexity in terms of both time and memory. 
This improvement is made possible by taking advantage of the
structural properties of kernels that can be expanded on a Fourier basis.
Furthermore, our approach lends itself well to parallel computation and can be efficiently implemented on modern GPU architectures. These advances are a significant step forward in both the theory and practice of kernel methods, paving the way for their routine application to large-scale datasets. A key component of our approach is the recent implementation of the non-uniform fast Fourier transform on GPU \citep[NUFFT,][]{shih2021nonuniform}.
The resulting computational efficiency enables the processing of tens of billions of data points in under a minute on a standard GPU, whereas conventional kernel approaches typically fail to scale beyond a few hundred thousand samples \citep{Meanti2020kernel, doumèche2024physicsinformedkernellearning}.

We illustrate our framework through three fundamental kernel learning tasks: Sobolev regression \citep{Nemirovski2000estimating}, physics-informed regression \citep{doumèche2025forecastingtimeseriesconstraints}, and additive model regression \citep{hastie1986generalized}. 
The ability to process massive datasets allows us to systematically assess the performance of the estimators, clearly highlighting the benefits of our approach. 
In particular, we present an additive model estimator based on kernels and Fourier expansions. Unlike classical spline-based additive models, our estimator exhibits more favorable scaling with the input dimension $d$ and introduces a new level of computational scalability. To support real-world applications, we provide the first GPU-based implementation of a kernel additive model that combines statistical accuracy with high computational efficiency. 
The code to reproduce our experiment is available at \href{https://github.com/NathanDoumeche/fast-kernel}{github.com/NathanDoumeche/fast-kernel}.


Together, our contributions form a unified framework of fast and theoretically sound kernel methods.
This framework offers substantial advantages for high-dimensional and large-scale learning tasks.


\section{Common setting}
\subsection{Nonparametric regression with kernel methods}
\paragraph{General assumptions.} 
Throughout this article, we adopt the standard nonparametric regression setting, in which the objective is to estimate an unknown function $f^\star$ from $n \in \mathbb{N^{\star}}$ independent and identically distributed observations $(X_1, Y_1), \dots, (X_n, Y_n)$ of a random pair $(X, Y)$, under the following assumptions:
\begin{itemize}
\item[(i)] The input variable $X$ takes values in an open domain $\Omega \subseteq \mathbb{R}^d$ for some $d \geq 1$, and the output $Y$ is real-valued;
\item[(ii)] The output is modeled as $Y = f^\star(X) + \varepsilon$,
where the noise term $\varepsilon$ satisfies $\mathbb{E}(\varepsilon \mid X) = 0$ and $\mathbb{E}(\varepsilon^2 \mid X) \leq \sigma^2$ for some $\sigma > 0$.
\end{itemize}

To guarantee that $f^\star$ is learnable from data, we impose the following regularity assumptions:
\begin{itemize}
\item[(iii)] The target function belongs to a Sobolev space of smoothness $s > d/2$, that is, $f^\star \in H^s(\Omega)$ for some known $s$;
\item[(iv)] The domain $\Omega$ is a bounded Lipschitz domain of $\mathbb{R}^d$ \citep[see, e.g.,][]{agranovich2015lispchitz}, included in the hypercube $[-L, L]^d$ for some $L > 0$.
\end{itemize}

Recall that $H^s(\Omega)$ denotes the Sobolev space of functions whose weak derivatives up to order $s$ belong to $L^2(\Omega)$. This space includes, for example, the Hölder space $C^s(\bar{\Omega})$ of continuously differentiable functions with bounded derivatives. The Lipschitz assumption allows the boundary of the domain to be non-differentiable,
encompassing a broad class of bounded sets, including smooth manifolds and common geometries, such as hypercubes.  

\paragraph{Kernel methods.} 

In kernel-based methods, this setting naturally leads to the minimization of the regularized empirical risk functional
\[
\mathcal{R}(f) = \frac{1}{n} \sum_{j=1}^n (f(X_j) - Y_j)^2 + \lambda \|f\|_{\mathcal{H}}^2,
\]
over a reproducing kernel Hilbert space (RKHS) $\mathcal{H} \subseteq H^s(\Omega)$, where $\lambda > 0$ is a regularization parameter. The norm $\|\cdot\|_{\mathcal{H}}$ serves as a regularizer and reflects the prior assumption that the true function $f^\star$ has a bounded RKHS norm, i.e., $\|f^\star\|_{\mathcal{H}} < \infty$. The resulting estimator
\[
\hat{f}_n = \arg\min_{f \in \mathcal{H}} \mathcal{R}(f)
\]
serves as an estimator of the target function $f^\star$.

A classical result in kernel theory states that the estimator $\hat{f}_n$ can be computed in closed form as
\[
\hat{f}_n(x) = \big(K(X_1, x), \dots, K(X_n, x)\big) \, (\mathbb{K}_n+\lambda I)^{-1} \mathbb{Y},
\]
where the function $K$ 
is the reproducing kernel associated with the RKHS $\mathcal{H}$, and $\mathbb{K}_n$ is the $n \times n$ kernel matrix with entries $(\mathbb{K}_n)_{j_1, j_2} = K(X_{j_1}, X_{j_2})$ \citep[Chapter~7]{bach2024learning}.
The main computational bottleneck of this approach lies in the analytical computation of the kernel function $K$ and in the inversion of the kernel matrix $\mathbb{K}_n$, which incurs a cost of $\mathcal{O}(n^3)$ in time and $\mathcal{O}(n^2)$ in memory.



\paragraph{Finite-dimensional approximation.}
To make the computation of the kernel estimator more tractable,
we consider a finite-dimensional approximation of $\mathcal{H}$ through a finite basis $\phi_1, \dots, \phi_D$  with $D\in \mathbb N^\star$ elements.
Thus, any function $f \in \mathcal{H}$ can be approximated by a function $f_\theta$ of the form
\begin{equation}
     f_\theta(x) = \theta_1 \, \bar{\phi}_1(x) + \dots + \theta_D \, \bar{\phi}_D(x) = \langle \phi(x), \theta \rangle,
     \label{eq:f_theta}
\end{equation}
where $\phi = (\phi_1, \dots, \phi_D)^\top$ denotes the truncated feature map, $\theta \in \mathbb{C}^D$ is the parameter vector, and the inner product is defined as $\langle x, y \rangle = \sum_{k=1}^D \bar{x}_k y_k$. 
In the present paper, we let $\phi$ be the truncated Fourier basis composed of trigonometric functions.

Since the RKHS approximation if finite-dimensional, there is a positive-definite matrix $M$ such that, for all function $f_\theta \in \mathcal{H}$ given by Equation~\ref{eq:f_theta}, we have  $\|f_\theta\|_{\mathcal{H}}^2 = \|M \theta\|_2^2$.
The empirical risk of $f_\theta$ then translates into an empirical risk $R$ on the parameter $\theta$:
\[
\mathcal R(f_\theta) := R(\theta) = n^{-1} \|\Phi \theta - \mathbb{Y}\|_2^2 + \lambda \|M \theta\|_2^2,
\]
where $\Phi = \begin{pmatrix} \phi(X_1) \mid \dots \mid \phi(X_n) \end{pmatrix}^\ast$ is the design matrix (with complex conjugate transpose), 
and $\mathbb{Y} = (Y_1, \dots, Y_n)^\top \in \mathbb{R}^n$ is the response vector. Minimizing this objective yields the parameter
\begin{equation}
    \hat{\theta} = \arg\min_{\theta \in \mathbb{C}^D} R(\theta) = (n^{-1}\Phi^*\Phi+\lambda M^\star M)^{-1} n^{-1}\Phi^*\mathbb Y,
    \label{eq:kenrel_reg}
\end{equation}
and the resulting estimator takes the form
\[
\hat{f}_n(x) = \langle \phi(x), \hat{\theta} \rangle.
\]

\subsection{Fourier expansion}


Working with a Fourier basis expansion $\phi$ allows us to precisely quantify the approximation error in the RKHS, as detailed below.

\paragraph{Fourier expansion on the Sobolev space $H^s(\Omega)$.}

The smoothness of a function in $H^s(\Omega)$ (endowed with any of the equivalent Sobolev norms of order $s$) is reflected in the decay rate of its Fourier coefficients. This fact is formalized by the following proposition.

\begin{prop}[Fourier series representation, {\citealp[][]{doumèche2024physicsinformedkernellearning}}]
Let $s \in \mathbb{N}^\star$. There exists a constant $C_\Omega > 0$ such that, for any function $f \in H^s(\Omega)$, there is a sequence of Fourier coefficients $\theta(f) \in \mathbb{C}^{\mathbb{Z}^d}$ such that:
\begin{itemize}
    \item[(i)] $f(x) = \langle \phi_\infty(x), \theta(f) \rangle$, where $\phi_\infty(x) = \left( \exp\left(-i \frac{\pi}{2L} \langle k, x \rangle \right) \right)_{k \in \mathbb{Z}^d}$;
    \item[(ii)] The following norm equivalence holds:
    \[
    \|f\|_{H^s(\Omega)}^2 \leq \sum_{k \in \mathbb{Z}^d} |\theta(f)_k|^2 \, (1 + \|k\|_2^{2s}) \leq C_\Omega \|f\|_{H^s(\Omega)}^2.
    \]
\end{itemize}
\label{prop:fourier_rep}
\end{prop}
Since the Fourier coefficient vector $\theta(f) \in \mathbb{C}^{\mathbb{Z}^d}$ is infinite-dimensional, it cannot be stored or manipulated directly. Thus, in line with the finite-dimensional approximation principle presented above, we truncate the Fourier expansion to a finite number of frequencies, indexed by a cutoff parameter $m \in \mathbb{N}^{\star}$. 
In other words,  
we consider the finite-dimensional approximation space
\[
H_{m} = \Big\{ f : x \mapsto \sum_{k_1 = -m}^{m} \cdots \sum_{k_d = -m}^{m} \theta_{k_1, \dots, k_d} \exp\big(i \frac{\pi}{2L} \langle k, x \rangle \big),\; \theta \in \mathbb{C}^{(2m+1)^d} \Big\},
\]
where $k = (k_1, \dots, k_d) \in \mathbb{Z}^d$ is a multi-index. The dimension of this kernel Hilbert space is thus $D = (2m+1)^d$, and the truncated feature map reads 
\[
\phi(x) = \Big( \exp\big(-i \frac{\pi}{2L} \langle k, x \rangle \big) \Big)_{\|k\|_\infty \leq m}.
\]
For any function $f \in H^s(\Omega)$, its truncated approximation $f_{m} \in H_{m}$ takes the form
\[
f_{m}(x) = \langle\phi(x), (\theta(f)_k)_{\|k\|_\infty \leq m}\rangle.
\]

\paragraph{Fourier truncation.} The truncation of the high-frequency Fourier modes is justified by the following approximation result.
Let $\mathbb{P}_X$ denote the marginal distribution of the input variable $X$ on the domain $\Omega$. From this point onward, we assume the following regularity condition on the marginal distribution $\mathbb{P}_X$:
\begin{itemize}
    \item[(v)] $\mathbb{P}_X$ is a probability measure on $\Omega$ that admits a density with respect to the Lebesgue measure, bounded above by a constant $\kappa > 0$, i.e., $\frac{d\mathbb{P}_X}{dx} \leq \kappa$.
\end{itemize}

\begin{prop}[Approximation of the Fourier expansion]
Under assumption (v), there exists a constant $C > 0$, depending only on $d$ and $\Omega$, such that, for all $n \in \mathbb{N}^\star$, all $f \in H^s(\Omega)$, and all truncation levels $m \in \mathbb N^\star$, the following inequality holds:
\[
\|f - f_{m}\|_{L^2(\mathbb{P}_X)}^2 \leq C \kappa \, \|f\|_{H^s(\Omega)}^2 \, m^{-2s}.
\]

\label{lem:fourier_approx}
\end{prop}
Under Assumptions (i) to (v), any estimator $\hat f$ of $f^\star$ has an error $\mathbb E( \|f^\star-\hat f\|_{L^2(\mathbb P_X)}^2)$ at least of the order of $n^{-2s/(2s+d)}$ \citep[e.g.,][Theorem~2.1]{tsybakov2009introduction}.
This convergence rate of $2s / (2s + d)$ is known as the Sobolev minimax rate.  
Letting $m = n^{1/(2s+d)}$ in Proposition~\ref{lem:fourier_approx} is thus sufficient to achieve a precision of $n^{-2s/(2s+d)}$. 
Therefore, since adding extra Fourier modes would not result in an improved asymptotical precision on $\hat f_n$, in what follows, we consider $m = n^{1/(2s+d)}$.

Note that, in Proposition~\ref{lem:fourier_approx} and throughout the remainder of the paper, we evaluate the quality of an estimator using the $L^2(\mathbb{P}_X)$-norm, defined as
$
\|f\|_{L^2(\mathbb{P}_X)}^2 = \mathbb{E}(f^2(X)).$
This is a standard metric in statistical learning theory \citep[see, e.g.,][]{steinwart2008, bach2024learning}, reflecting the fundamental principle that one can only hope to estimate $f^\star$ accurately in regions where data are observed.

To sum up, in what follows, we consider the estimator~\ref{eq:kenrel_reg} on $H_m$, built on a truncated Fourier basis of level $m = n^{1/(2s+d)}$, with a matrix $M$ encoding the choice of the kernel norm.

\subsection{Fast optimization}
\label{sec:optim}

\paragraph{Naive implementation.}
Since the total number of Fourier modes is $m^d = n^{d/(2s + d)}$, a naive implementation of Equation~\ref{eq:kenrel_reg} would incur a computational complexity of $\mathcal{O}(n^{1 + 2d/(2s + d)})$, due to the cost of forming the matrix product $\Phi^\ast \Phi$. 
Though this is already more efficient than the kernel approach in $\mathcal O(n^3)$, 
we show in this section how to further reduce this complexity to $\mathcal{O}(n \log n)$.
This is nearly optimal, since even reading the full dataset incurs a lower bound of $\mathcal{O}(n)$. 
Furthermore, using a complex exponential basis to discretize the kernel regression problem is further justified by a set of computational techniques (fast transforms) that allow the estimator $\hat{\theta}$ to be computed significantly faster than with the naive approach.

\paragraph{Covariance vector.} A naive computation of the covariance 
vector $v = \Phi^\ast \mathbb{Y} / n$ via explicit matrix–vector multiplication leads to a complexity of $\mathcal{O}(n^{1 + d/(2s + d)})$. However, each component of $v$ can be written explicitly as
\[
v_k = \frac{1}{n} \sum_{j=1}^n Y_j \exp\big(i \pi \langle k, \frac{X_j}{2L}\rangle \big),
\]
which corresponds to the type-I non-uniform Fourier transform (NUFFT) of the weighted signal $\mathbb{Y}$ with respect to the non-uniform spatial locations  $(X_1, \hdots, X_n)$. 
Efficient NUFFT algorithms have been developed to compute this transform with time and memory complexity $\mathcal{O}(n \log n)$.
 Moreover, these algorithms can be parallelized on modern GPUs, benefiting from recent advances in hardware-optimized fast summation techniques. In what follows, we rely on the \texttt{cuFINUFFT} library for the GPU-based implementation of the NUFFT in dimensions 1, 2, and 3 \citep{shih2021nonuniform}.

\paragraph{Covariance matrix.} A naive computation of the empirical covariance matrix $\hat \Sigma= \Phi^\ast \Phi / n$ via explicit matrix–matrix multiplication incurs a computational complexity of $\mathcal{O}(n^{1 + 2d/(2s + d)})$. However, one can observe that $\hat \Sigma$ is both Hermitian and $d$-level block Toeplitz. 
Specifically, it satisfies $\hat \Sigma^\ast = \hat \Sigma$ and, for all $k_1, k_2 \in \{-m, \dots, m\}^d$, the relation
$\hat \Sigma_{k_1, k_2} = \hat \Sigma_{0, k_2 - k_1}$.
This means that the entire matrix $\hat \Sigma$ is fully determined by its first row, i.e., the $(2m + 1)^d$ values $(\hat \Sigma_{0, k})_{\|k\|_\infty \leq m}$.
Each entry can be expressed as
\[
\hat \Sigma_{0, k} = \frac{1}{n} \sum_{j=1}^n \exp\big(i \pi \langle k, \frac{X_j}{2L} \rangle \big),
\]
which corresponds to the type-I non-uniform Fourier transform of the constant signal $(1, \dots, 1)$ evaluated at the sample points $(X_1, \hdots, X_n)$. These values can be computed in time and memory complexity $\mathcal{O}(n \log n)$ using NUFFT algorithms with GPU acceleration. Furthermore, for any $x \in \mathbb{C}^{(2m + 1)^d}$, the matrix–vector product $\hat \Sigma x$ can be performed efficiently via FFT in $\mathcal{O}(m^d \log m) = \mathcal{O}(n^{d/(2s + d)} \log n)$ time; see \citet[Theorem~4.8.2,][]{golub2013matrix} for the one-dimensional case and the multidimensional construction of \citet{lee1986fast}.

\paragraph{Matrix inversion.} 

In order to speed up the matrix inversion, we assume that for any $x \in \mathbb{C}^{(2m + 1)^d}$, the matrix–vector product $M^\star Mx$ can be performed in 
$\mathcal{O}(D \log D) = \mathcal{O}(n^{d/(2s + d)} \log n)$. This condition is satisfied if $M$ is diagonal or, as shown above, if $M^\star M$ is a covariance matrix.
Then, since the matrix $\hat \Sigma$ is Hermitian and positive semi-definite,  the matrix–vector product $\hat \Sigma x$ can also be performed in $\mathcal{O}(n^{d/(2s + d)} \log n)$ time via FFT.
All in all, linear systems involving $\hat \Sigma + \lambda M^\star M$ can thus be solved efficiently using the conjugate gradient method. 
The total complexity of solving such systems is reduced to $\mathcal{O}(n^{2d/(2s + d)} \log n) = o(n)$, which is sublinear in the number of data points.

\paragraph{Conclusion.} The estimator $\hat \theta$ in Equation~\ref{eq:kenrel_reg} can be computed with both time and memory complexity of $\mathcal{O}(n \log n)$. Furthermore, this computation is highly parallelizable and benefits from recent advances in GPU hardware. This makes it well-suited for large-scale kernel regression tasks.

\section{Sobolev regression}
\label{sec:sob}
\subsection{Sobolev kernel regression}
\label{sec:sob_1}
In this section, we study the Sobolev kernel regression where $\mathcal H = H^s(\Omega)$ and the RKHS norm is given by $\|f\|_{\mathcal H}^2 = \sum_{k \in \mathbb{Z}^d} |\theta(f)_k|^2 \, (1 + \|k\|_2^{2s})$. 
 (Since all Sobolev norms are equivalent, the choice of the Sobolev norm does not impact the rate of convergence of the kernel estimator.)
 In the finite-dimensional approximation using $H_m$, one has
 $\|f_\theta \|_{\mathcal H}^2 = \|S\theta\|_2^2$, where the matrix $S\in \mathbb{C}^{(2m+1)^d\times (2m+1)^d}$ is diagonal with entries $S_{k_1,k_2}=\sqrt{1+\|k\|^{2s}_2}\mathbf{1}_{k_1=k_2}$. 
 Using tools from kernel theory \citep{mourtada2022an, bach2024learning}, it is then possible to derive the convergence rate of the estimator $\hat{\theta}$ as a function of the sample size $n$. 
 This result is encapsulated in the following proposition.

\begin{prop}[Sobolev kernel regression]
    Let $\hat{\theta}$ be the Sobolev kernel estimator given by Equation~\ref{eq:kenrel_reg} with $M = S$.
    Then,
    \begin{align*}
        \hat \theta &= \mathrm{argmin}_{\theta \in \mathbb C^{(2m+1)^d}} n^{-1}\|\Phi\theta-\mathbb Y\|_2^2 + \lambda \|S\theta\|_2^2
        \\
        &= (n^{-1}\Phi^*\Phi+\lambda S^2)^{-1} n^{-1}\Phi^*\mathbb Y.
    \end{align*}
    Under Assumptions (i)-(v), 
\begin{align*}
    \mathbb E(\|\hat f_{n}-f^\star\|_{L^2(\mathbb P_X)}^2) &\leq \inf_{\theta \in \mathbb C^{(2m+1)^d}}\Big\{\big(2+6\big(1+\frac{\alpha^2}{\lambda n}\big)^2\big)\mathbb E(\|f^{\star}-f_{\theta}\|_{L^2(\mathbb P_X)}^2) + \lambda \big(1+\frac{\alpha^2}{\lambda n}\big)^2 \|S\theta\|_2^2 \Big\}\\
    &\qquad + \sigma^2 n^{-1} \big(1+\frac{\alpha^2}{\lambda n}\big) (2m+1)^d,
\end{align*}
    where  $\alpha = \sum_{k\in \mathbb Z^d} \frac{1}{1+\|k\|^{2s}_2} < \infty$.
Therefore, choosing the regularization parameter in the range $n^{-1} \leq \lambda \leq n^{-2s/(2s + d)}$ and setting the truncation level as $m = n^{1/(2s + d)}$ yields the convergence rate
\[
\mathbb{E}(\|f_{\hat{\theta}} - f^\star\|_{L^2(\mathbb{P}_X)}^2) = \mathcal O\big(n^{-2s/(2s + d)}\big).
\]   
    \label{prop:sob_ini}
\end{prop}

This convergence rate of $2s / (2s + d)$ matches the Sobolev minimax rate, meaning that it is the fastest rate achievable under Assumptions (i) to (v).
The existence of minimax Sobolev estimators is well-known in the statistical learning literature \citep[e.g.,][Chapter~2]{Nemirovski2000estimating}.
However, Proposition~\ref{prop:sob_ini} establishes a clear connection between kernel methods and Fourier analysis. More importantly, since $S$ is diagonal, this estimator can be computed exactly on a GPU with complexity $\mathcal{O}(n \log n)$, as shown in Section~\ref{sec:optim}.
This property enables kernel methods to scale efficiently to large datasets.

\paragraph{One dimensional example.} \begin{wrapfigure}[12]{r}{0.35\textwidth}
    \vspace{-1em}
    \includegraphics[width=\linewidth]{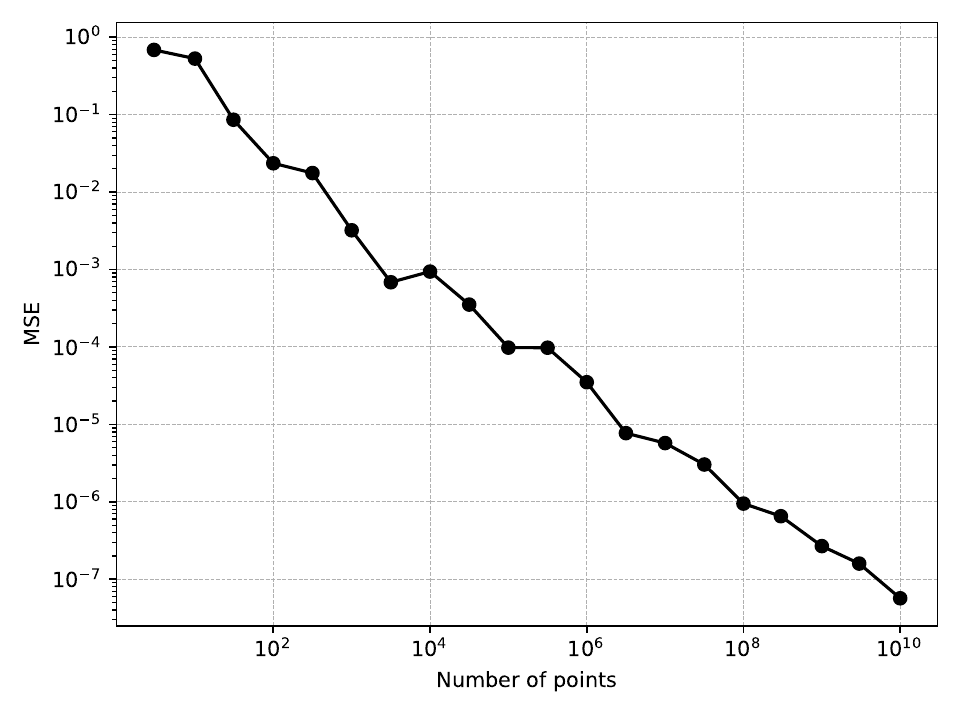}
    \caption{One-dimensional Sobolev kernel regression}
    \label{fig:sob1d}
\end{wrapfigure}
The goal of this experiment is to demonstrate the ability of our method to efficiently handle datasets containing tens of billions of samples, while experimentally recovering the theoretical convergence rate. 
We consider the following simple one-dimensional regression setting:
\begin{itemize}
    \item The input variable $X$ is uniformly distributed over the interval $\Omega = ]0, 1[$;
    \item The output is given by $Y = \exp(X) + \varepsilon$, where the noise $\varepsilon \sim \mathcal{N}(0, 1)$ is independent of $X$.
\end{itemize}
We run the Sobolev kernel estimator with smoothness parameter $s = 1$, regularization $\lambda = n^{-2/3}$, and truncation level $m = n^{1/3}$. The algorithm successfully processes $n = 10^{10}$ data points in approximately one minute on a standard GPU (NVIDIA T4), using complex-128 precision. 
Figure~\ref{fig:sob1d} reports the test error $\mathbb{E}(\|f_{\hat{\theta}} - f^\star\|_{L^2(\mathbb{P}_X)}^2)$, estimated over a separate test set of $10^4$ samples, and averaged over 20 resamples. The observed performance closely matches the theoretical Sobolev minimax rate of $n^{-2/3}$, providing strong empirical evidence of the method's numerical stability and scalability.

\subsection{Low-bias Sobolev kernel regression}

\paragraph{Dependency on the smoothness $s$.} 
If the target function $f^\star$ is known to have smoothness $s$, then it also belongs to all Sobolev spaces $H^{s_2}(\Omega)$ for any $s_2 < s$. As a result, Proposition~\ref{prop:sob_ini} guarantees that all Sobolev kernel estimators with smoothness parameter $s_2 < s$ are well-defined and provably converge to $f^\star$. From a theoretical point of view, higher values of $s$ yield faster convergence rates.
However, the non-asymptotic bound in Proposition~\ref{prop:sob_ini} also reveals a bias term involving $\lambda \|S \theta\|_2^2 \geq \lambda \|f^\star\|_{H^s(\Omega)}^2$.  
Overall, this term scales as $n^{-2s/(2s+d)} \|f^\star\|_{H^s(\Omega)}^2$, highlighting the classical trade-off inherent in the choice of $s$: larger values improve the convergence rate but come at the cost of increased dependence on the Sobolev norm of the target function.

For example, consider the function $f(x) = x^k$ on the interval $]-1, 1[$. 
One can compute
\[
\|f\|_{H^1(]-1, 1[)}^2 = \int_{-1}^1 f(x)^2 + f'(x)^2 \, dx = \frac{2}{2k+1} + \frac{2k^2}{2k - 1},
\]
whereas
\[
\|f\|_{H^k(]-1, 1[)}^2 \geq \int_{-1}^1 \big(f^{(k)}(x)\big)^2 \, dx = 2(k!)^2,
\]
showing that $\|f\|_{H^k(]-1, 1[)}^2 \gg \|f\|_{H^1(]-1, 1[)}^2$ for large $k$. This illustrates how higher-order Sobolev norms can grow rapidly, even for smooth functions, and lead to substantial regularization bias despite improved asymptotic rates.

To mitigate the increasing regularization bias associated with large values of $s$, we propose an alternative estimator that penalizes the unweighted $\ell_2$ norm of the parameter vector, i.e., $\lambda \|\theta\|_2^2$, instead of the Sobolev-weighted norm $\lambda \|S \theta\|_2^2$. 
\begin{prop}[Low-bias Sobolev kernel regression] 
Let $\hat{\theta}$ be the low-bias Sobolev kernel estimator given by Equation~\ref{eq:kenrel_reg} with $M = I$, where $I$ is the identity matrix.
Then,
    \begin{align*}
        \hat \theta & = \mathrm{argmin}_{\theta \in \mathbb C^{(2m+1)^d}} n^{-1}\|\Phi\theta-\mathbb Y\|_2^2 + \lambda \|\theta\|_2^2
        \\
        &= (n^{-1}\Phi^*\Phi+\lambda I)^{-1} n^{-1}\Phi^*\mathbb Y.
    \end{align*}
    Under Assumptions (i)-(v), 
\begin{align*}
    \mathbb E(\|\hat f_{n}-f^\star\|_{L^2(\mathbb P_X)}^2) &\leq \inf_{\theta \in \mathbb C^{(2m+1)^d}}\Big\{\big(2+6(1+r)^2\big)\mathbb E(\|f^{\star}-f_{\theta}\|_{L^2(\mathbb P_X)}^2) + \lambda (1+r)^2 \|\theta\|_2^2 \Big\}\\
    &\qquad + \lambda \sigma^2 r(1+r),
\end{align*}
    where $r=\frac{(2m+1)^d}{\lambda n}$.
    Therefore, choosing the regularization parameter as $\lambda = n^{-2s/(2s+d)}$ and setting the truncation level as $m = n^{1/(2s+d)}$ yields the convergence rate
\[\mathbb E[\|f_{\hat \theta}-f^\star\|_{L^2(\mathbb P_X)}^2] = \mathcal{O}(n^{-2s/(2s+d)}).
\] 
    \label{prop:sob_rest}
\end{prop}

As before, this estimator can be computed exactly on a GPU with complexity $\mathcal{O}(n \log n)$.

\paragraph{Dependency in $s$ in the one-dimensional case.}
The goal of this experiment is to compare the performance of the Sobolev kernel estimator with that of its low-bias variant. We consider the following one-dimensional regression setting:
\begin{itemize}
    \item[(i)] The input variable $X$ is uniformly distributed over the interval $\Omega = ]0, 1[$;
    \item[(ii)]The response is given by $Y = 25\,|X - 0.5|^3 + \varepsilon$, where the noise term $\varepsilon \sim \mathcal{N}(0, 1)$ is independent of $X$.
\end{itemize}
We implement both kernel estimators across multiple sample sizes $n$, and evaluate their MSE on a fixed test set of $10^4$ points. To ensure a meaningful comparison, we use the same regularization and truncation schedules for both methods, setting $\lambda_n = n^{-2s/(2s+1)}$ and $m = n^{1/(2s+1)}$ for each smoothness parameter $s$. We consider $40$ values of $s$ uniformly spaced between the minimal admissible value $s = d/2 = 1/2$ and $s = 10$.

\begin{figure}
    \centering
    \includegraphics[width=0.45\linewidth]{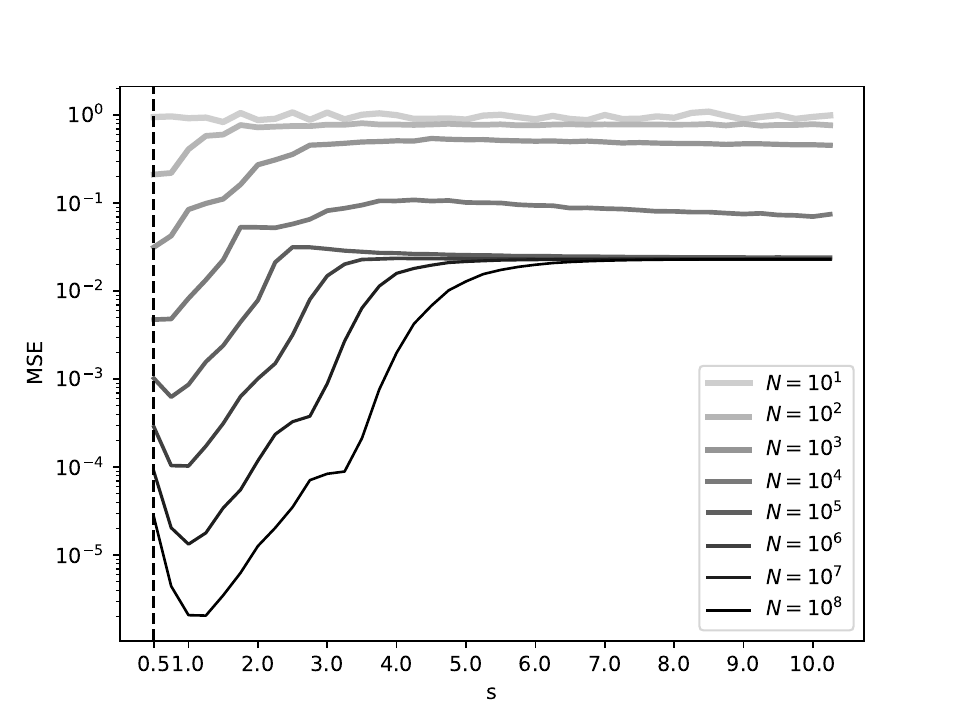}\includegraphics[width=0.45\linewidth]{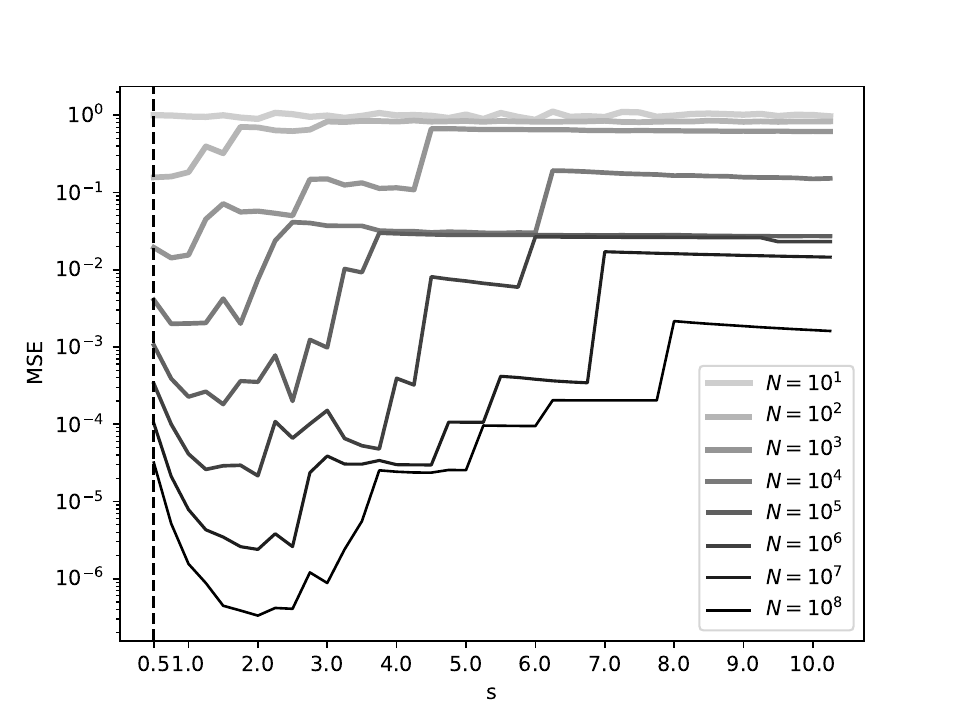}
    \caption{Sobolev regression (Left), and low-bias regression (Right) in dimension $d=1$ 
    }
    \label{fig:s-comparison}
\end{figure}

Figure~\ref{fig:s-comparison} reports the MSE of the two kernel estimators, averaged over $10$ independent resamples for each value of $s$. This large-scale experiment involves training $400$ kernel estimators at sample size $n = 10^8$, and completes in approximately 5 minutes on a standard GPU, illustrating the computational efficiency of both kernel methods.

As expected, for any fixed value of $s$, the error of both kernel estimators decreases as the number of observations $n$ increases. In addition, we clearly observe the trade-off discussed earlier regarding the choice of the smoothness parameter $s$. Indeed, the target function $f^\star(x) = 25\,|x - 0.5|^3$ belongs to the Sobolev space $H^3([0,1])$ but not to $H^4([0,1])$, which implies that $s = 3$ is the largest integer smoothness level for which the minimax convergence rate remains theoretically valid.
However, in practice, the value $s = 3$ does not correspond to the hyperparameter that yields the lowest MSE. Interestingly, in both kernel estimators, the optimal choice of $s$ increases gradually with the sample size $n$. This empirical behavior is consistent with the theoretical results of Propositions~\ref{prop:sob_ini} and~\ref{prop:sob_rest}, which indicate that the optimal $s$ should asymptotically converge to $3$ as $n$ becomes large.

In practice, the low-bias Sobolev kernel consistently outperforms the Sobolev kernel. For example, at $n = 10^8$, its minimum MSE, achieved at $s = 2$, is approximately ten times lower than that of the Sobolev kernel, whose minimum occurs at $s = 1$.  Additionally, the shift toward higher values of $s$ occurs more rapidly for the low-bias kernel. This is due to its reduced sensitivity to the explosion of the Sobolev norm as $s$ increases, allowing it to benefit from the improved theoretical convergence rates associated with higher regularity at smaller sample sizes.

\paragraph{Dependency in $s$ in the two-dimensional case.}
In this experiment, we aim to compare the performance of the Sobolev kernels in dimension $2$, for the task of approximating smooth functions. The regression setting is defined as follows:
\begin{itemize}
    \item[(i)] The input $X=(X^{(1)}, X^{(2)})$ is uniformly distributed over the square domain $\Omega = ]0, 1[^2$;
    \item[(ii)] The response is given by $Y = \exp(X^{(1)})\cos(X^{(2)}) + \varepsilon$, where the noise term $\varepsilon \sim \mathcal{N}(0, 1)$ is independent of $X$.
\end{itemize}

\begin{figure}
    \centering
    \includegraphics[width=0.45\linewidth]{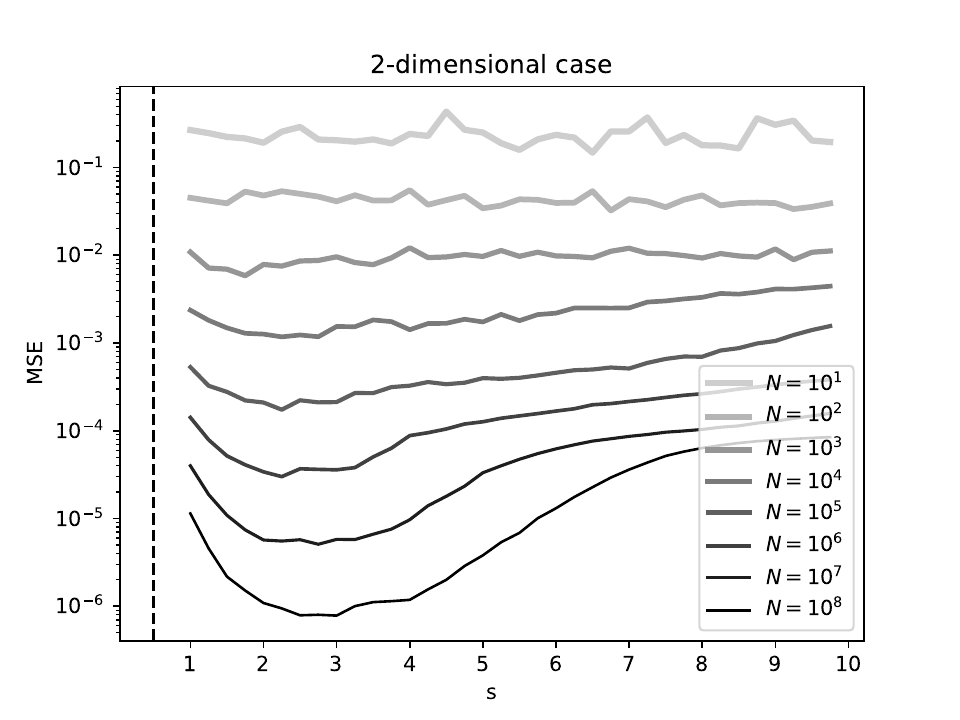}
    \includegraphics[width=0.45\linewidth]{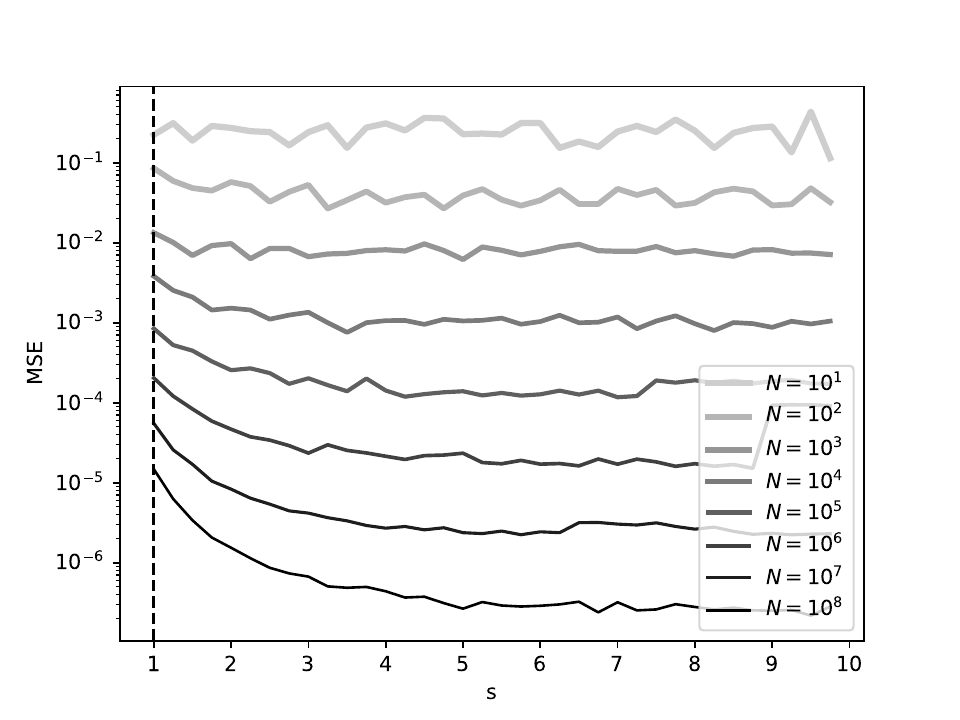}
    \caption{Sobolev regression (Left), and low-bias regression (Right) in dimension $d=2$}
    \label{fig:s-comparison-2d}
\end{figure}

We implement both kernel estimators for several sample sizes $n$, and evaluate their MSE on a test set of $10^4$ points. To ensure a meaningful comparison, we adopt the same regularization and truncation schedules for both methods, setting $\lambda_n = n^{-2s/(2s + 2)}$ and $m = n^{1/(2s + 2)}$. We consider $40$ values of the smoothness parameter $s$, uniformly spaced between $s = d/2 = 1$ and $s = 10$. Figure~\ref{fig:s-comparison-2d} reports the MSEs of both techniques, averaged over $10$ independent resamples for each value of $s$. This large-scale experiment, which involves training $400$ kernel estimators at $n = 10^8$, completes in approximately 21 minutes on a standard GPU (NVIDIA L4).

Once again, the low-bias Sobolev kernel consistently outperforms the Sobolev kernel. In this example, the target function satisfies $f^\star \in \cap_{s \in \mathbb{N}} H^s(\Omega)$, meaning that it is infinitely smooth and the theoretically optimal regularity parameter is $s^\star = \infty$. As in the one-dimensional case, the optimal value of $s$ for the Sobolev kernel increases more slowly toward $s^\star$ than for the low-bias estimator, which benefits more rapidly from the higher smoothness of $f^\star$.

\section{Physics-informed regression}
Interestingly, our approach, which combines Fourier analysis with kernel methods, applies naturally to the framework of physics-informed machine learning (PIML). PIML incorporates prior physical knowledge--typically expressed as a partial differential equation (PDE)--into regression problems in order to regularize the empirical risk \citep{karniadakis2021piml}. 
Therefore, in this context, we assume that
\begin{itemize}
    \item[(vi)] There is a known linear differential operator $\mathcal{D}$  with constant coefficients such that $\forall x\in \Omega,\;\mathcal{D}(f^\star,x) = 0$. 
\end{itemize}
Recall that 
$\mathcal{D}$ is linear with constant coefficients if
\[
\mathcal{D}(f, x) = \sum_{|\alpha| \leq s} a_\alpha \, \partial^\alpha f(x), \quad \text{with } a_\alpha \in \mathbb{R}.
\]
It turns out that this regression setting can be formulated as a kernel method on $\mathcal{H} = H^s(\Omega)$ by minimizing the empirical risk
$n^{-1}\sum_{j=1}^n(f_\theta(X_j)-Y_j)^2 + \lambda \|f_\theta\|_{H^s(\Omega)}^2 + \mu \int_\Omega \mathcal{D}(f_\theta, x)^2dx$, as studied in \citet{doumeche2024physicsinformed} and \citet{ doumèche2024physicsinformedkernellearning}.
In what follows, we show that this PDE-based penalty can be efficiently integrated into our Fourier kernel framework, where it naturally benefits from NUFFT-based acceleration.
\paragraph{Tractable domains.}
When the domain $\Omega$ is regular enough to allow for the analytical computation of its Fourier transform, 
the regression problem can be formulated equivalently on the finite-dimensional space $H_m$ as a kernel estimator, of the form:
\begin{align*}
        \hat \theta &= \hbox{argmin}_{\theta \in \mathbb C^{(2m+1)^d}} n^{-1}\sum_{j=1}^n(f_\theta(X_j)-Y_j)^2 + \lambda \|f_\theta\|_{H^s}^2 + \mu \int_\Omega \mathcal{D}(f_\theta, x)^2dx \\
        &= \hbox{argmin}_{\theta \in \mathbb C^{(2m+1)^d}} n^{-1}\|\Phi\theta-\mathbb Y\|_2^2 + \lambda \|S\theta\|_2^2 + \mu \|C^{1/2}D\theta\|_2^2
        \\
        &= (n^{-1}\Phi^*\Phi+\lambda S^2+\mu D^\star CD)^{-1} n^{-1}\Phi^*\mathbb Y,
    \end{align*}
    where $S$ is the Sobolev matrix defined in Section~\ref{sec:sob_1},  the Fourier matrix $C$ of $\Omega$ is such that, for all $k_1, k_2 \in \mathbb C^{(2m+1)^d}$,
    \[
    C_{k_1, k_2} = (4L)^{-d} \int_{\Omega} \exp(i\langle k_1-k_2, x\rangle \pi/(2L))dx,
    \]
    and $D$ is the $(2m+1)^d\times (2m+1)^d$ diagonal matrix such that 
    \[D_{k,k} =  \sum_{|\alpha| \leq s} a_\alpha \Big(\frac{-i\pi}{2L}\Big)^{|\alpha|}\prod_{\ell =1}^d  (k_\ell)^{\alpha_\ell}.
    \]
Since $D$ is a diagonal matrix and $C$ is a $d$-level block Toeplitz and Hermitian matrix, the matrix-vector product involving $D^\ast C D$ can be performed in $\mathcal{O}(m^d \log m)$ operations. As a result, the conjugate gradient inversion of the full system matrix $(n^{-1} \Phi^\ast \Phi + \lambda S^2 + \mu D^\ast C D)$ can be carried out with overall complexity $\mathcal{O}(n \log n)$.

\paragraph{Untractable domains.} 
When the Fourier decomposition of the domain $\Omega$ is not available in closed form, a common strategy in physics-informed machine learning consists in sampling a set of $n_r \in \mathbb{N}^\star$ collocation points $(X^{(r)}_1, \dots, X^{(r)}_{n_r})$ from a known distribution $\mathbb{P}_{X^{(r)}}$ over $\Omega$. Based on these points, we then define the following kernel estimator:
\begin{align*}
        \hat \theta &= \hbox{argmin}_{\theta \in \mathbb C^{(2m+1)^d}} n^{-1}\sum_{j=1}^n(f_\theta(X_j)-Y_j)^2 + \lambda \|f_\theta\|_{H^s}^2 + \mu n_r^{-1} \sum_{\ell=1}^{n_r}\mathcal{D}(f_\theta, X^{(r)}_j)^2 \\
        &= \hbox{argmin}_{\theta \in \mathbb C^{(2m+1)^d}} n^{-1}\|\Phi\theta-\mathbb Y\|_2^2 + \lambda \|S\theta\|_2^2 + \mu n_r^{-1} \|\Phi D\theta\|_2^2
        \\
        &= (n^{-1}\Phi^*\Phi+\lambda S^2+\mu n_r^{-1}D^\star (\Phi^{(r)})^\star \Phi^{(r)} D)^{-1} n^{-1}\Phi^*\mathbb Y,
    \end{align*}
    where $\Phi^{(r)} = ( \phi(X_1^{(r)}) \mid \hdots \mid \phi(X_n^{(r)}))^\star$.
Since $D$ is a diagonal matrix and $(\Phi^{(r)})^\ast \Phi^{(r)}$ is a $d$-level block Toeplitz and Hermitian matrix, the associated matrix–vector products can be computed in $\mathcal{O}(m^d \log m)$ operations. Consequently, the conjugate gradient inversion of the system
\[
n^{-1} \Phi^\ast \Phi + \lambda S^2 + \mu n_r^{-1} D^\ast (\Phi^{(r)})^\ast \Phi^{(r)} D
\]
can be performed with overall complexity $\mathcal{O}(n \log n)$.

\begin{figure}
    \centering
    \includegraphics[width=0.45\linewidth]{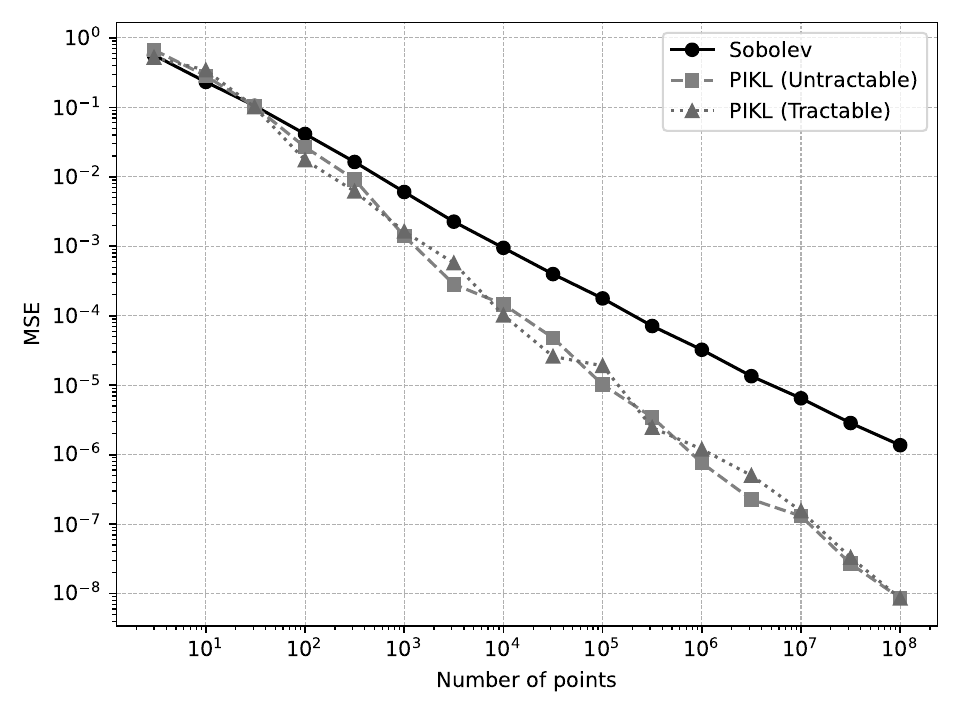}
    \includegraphics[width=0.45\linewidth]{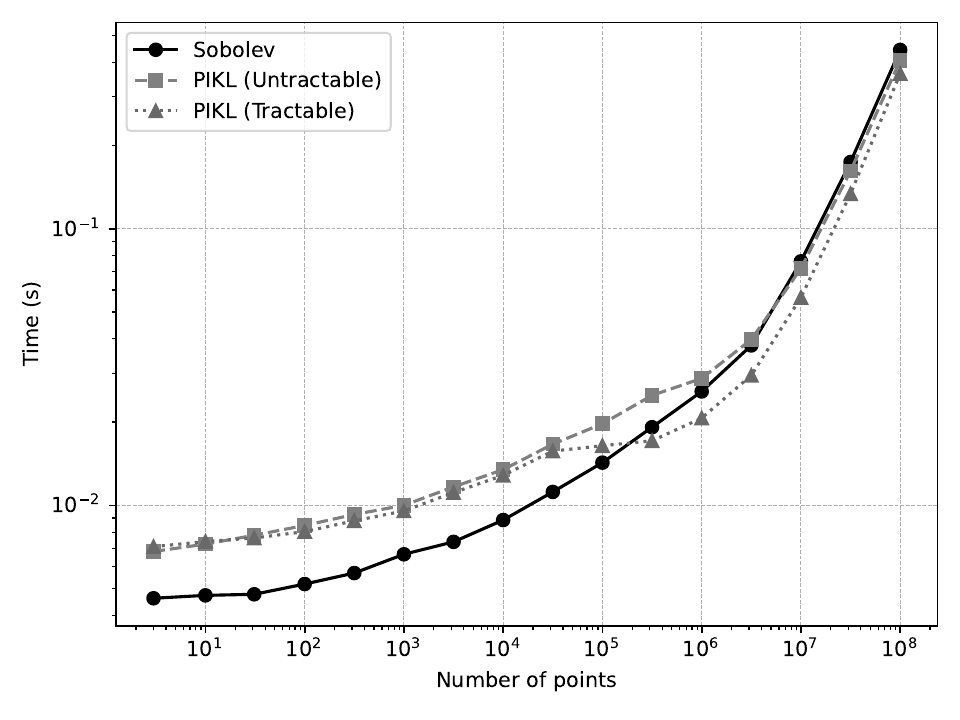}
    \caption{Physics-informed kernel regression (averaged over 20 samples)
    }
    \label{fig:pik1d}
\end{figure}

\paragraph{One dimensional example.}
In this experiment, we consider the following one-dimensional regression setting:
\begin{itemize}
    \item[(i)] The input variable $X$ is uniformly distributed over the interval $\Omega = ]0, 1[ \subseteq [-\pi/2, \pi/2]$, so that $L = \pi/2$;
    \item[(ii)]  The response is given by $Y = \exp(X) + \varepsilon$, where $\varepsilon \sim \mathcal{N}(0, 1)$ is independent of $X$;
    \item[(iii)]  The prior knowledge takes the form of a first-order linear differential constraint: $(f^\star)' - f^\star = 0$;
    \item[(iv)] The parameters are set as $s = d = 1$, $m = n^{1/(2s + d)} = n^{1/3}$, $\lambda = n^{-2s/(2s + d)} = n^{-2/3}$, and $\mu = 1$.
\end{itemize}
All the kernel methods process $n = 10^8$ samples in under only one second on a standard GPU (NVIDIA L4). Figure~\ref{fig:pik1d} reports the test error $\mathbb{E}(\|f_{\hat{\theta}} - f^\star\|_{L^2([0, 1])}^2)$, evaluated on a test set of $10^4$ samples.
We see that adding physics improves the performance of the estimator, while taking approximately the same running time.

\section{Additive models}
In addition to differential constraints, our Fourier–NUFFT framework can also accommodate linear constraints on the shape of the regression function. In this section, we show how to efficiently incorporate additivity constraints, yielding a GPU-compatible algorithm with an overall complexity of $\mathcal{O}(n \log n)$.


\paragraph{Curse of the dimension.}
A well-known limitation of many machine learning methods is that their performance deteriorates as the dimensionality $d$ of the input space increases. This phenomenon is commonly referred to as the curse of dimensionality.
This effect is particularly evident for the Sobolev kernel estimators introduced in Section~\ref{sec:sob}, since the corresponding minimax rate $\mathcal{O}(n^{-2s/(2s + d)})$ implies that the number of required samples grows exponentially with $d$.  Specifically, achieving a target accuracy of $\delta$ requires at least $n = \delta^{-1 - d / (2s)}$ observations.

\paragraph{Additive model.} 
To mitigate this phenomenon, additional structural assumptions are required. One common and straightforward approach is to exclude interaction effects between input variables, resulting in the following additive model:
\begin{itemize}
    \item[(vii)] One has $f^\star(x_1, \dots, x_d) = g^\star_1(x_1)+\dots +g_d^\star(x_d)$, where $g^\star_1$, $\dots$, $g^\star_d$ are univariate functions.  
\end{itemize}
The smoothness assumption on $f^\star$ naturally transfers to the component functions $g^\star_\ell$. 
In particular $f^\star$ lies in a Sobolev space of order $s$, if and only if each function $g^\star_\ell$ belongs in a Sobolev space of order $s$.
\begin{itemize}
    \item[(iii)'] There exists a known smoothness parameter $s > 1/2$ such that each component function $g^\star_1, \dots, g^\star_d$ belongs to the univariate Sobolev space of order $s$.
\end{itemize}
The low-bias Sobolev kernel can thus be naturally extended to the additive model setting. In this case, the regression function takes the form
\begin{equation*}
    f_\theta(x_1, \dots, x_d) = \langle \phi(x_1), \theta_1 \rangle + \dots + \langle \phi(x_d), \theta_d \rangle,
    \label{eq:param_ma}
\end{equation*}
and is parametrized by a concatenated vector $\theta = (\theta_1, \dots, \theta_d) \in \mathbb{C}^{d(2m + 1)}$.

\begin{prop}[Low-bias additive kernel regression]
Let $\hat \theta$ be the truncated Fourier estimator 
defined by
    \begin{align*}
        \hat \theta &= \mathrm{argmin}_{ \theta \in \mathbb C^{d(2m+1)}} n^{-1}\|\Phi_1\theta_1+\dots +\Phi_d \theta_d-\mathbb Y\|_2^2 + \lambda \|\theta\|_2^2
        \\
        &= (\hat \Sigma+\lambda I)^{-1} \begin{pmatrix}
            n^{-1}\Phi_1^*\mathbb Y\\
            \vdots\\
            n^{-1}\Phi_d^\star \mathbb Y
        \end{pmatrix},
    \end{align*}
    where $$\Phi_\ell = \begin{pmatrix} \phi(X_{1, \ell}) \mid \cdots \mid \phi(X_{n,\ell})\end{pmatrix}^\star,$$  $I$ is the $d(2m + 1)\times d(2m + 1)$ identity matrix, and $M$ is the block-matrix such that the $(\ell_1,\ell_2)$-block is $n^{-1}\Phi_{\ell_1}^*\Phi_{\ell_2}$, i.e., 
    $$\hat \Sigma = \begin{pmatrix}
            n^{-1}\Phi_1^*\Phi_1 & \hdots & n^{-1}\Phi_1^*\Phi_d\\
            \vdots & \ddots & \vdots\\
            n^{-1}\Phi_d^*\Phi_1& \hdots & n^{-1}\Phi_d^*\Phi_d
        \end{pmatrix}.$$
    Then, under Assumptions (i), (ii), (iii)', (iv), (v), and (vii), one has
\begin{align*}
    \mathbb E(\|\hat f_{n}-f^\star\|_{L^2(\mathbb P_X)}^2) &\leq \inf_{\theta \in \mathbb C^{d(2m+1)}}\Big(\big(2+6(1+r)^2\big)\mathbb E(\|f^{\star}-f_{\theta}\|_{L^2(\mathbb P_X)}^2) + \lambda (1+r)^2 \|\theta\|_2^2 \Big)\\
    &\qquad + \lambda \sigma^2 r(1+r),
\end{align*}
    where 
    $r=\frac{d(2m+1)}{\lambda n}$.
    Therefore, choosing the regularization parameter as $\lambda = n^{-2s/(2s+1)}$ and setting the truncation level as $m = n^{1/(2s+1)}/d$ yields the convergence rate
    \[
    \mathbb E(\|f_{\hat \theta}-f^\star\|_{L^2(\mathbb P_X)}^2) = \mathcal{O}( n^{-2s/(2s+1)}).
    \]
    \label{prop:additive}
\end{prop}
Notice that the additive assumption allows us to recover the more favorable univariate minimax rate of $n^{-2s/(2s + 1)}$, as in the case $d = 1$. The scaling $m = n^{1/(2s + 1)} / d$ is consistent with existing results on generalized additive models (GAMs) in high dimensions \citep[Section~3.3,][]{wood2014generalized}.

\paragraph{Complexity.}

As explained in Section~\ref{sec:optim}, the terms $n^{-1} \Phi_1^\ast \mathbb{Y}$ and the diagonal blocks $n^{-1} \Phi_\ell^\ast \Phi_\ell$ of the matrix $\hat \Sigma$ can be efficiently computed using the NUFFT. As a result, the vector
$(\Phi_1^\ast \mathbb{Y},\dots,
n^{-1} \Phi_d^\ast \mathbb{Y})^\top$ and the diagonal of $\hat \Sigma$
can be computed with overall complexity $\mathcal{O}(dn \log n)$. Moreover, the off-diagonal block $n^{-1} \Phi_{\ell_1}^\ast \Phi_{\ell_2}$ corresponds to the 2d-NUFFT of the constant function equal to $1$, evaluated at the points $(X_{1, \ell_1}, -X_{1, \ell_2}), \dots, (X_{n, \ell_1}, -X_{n, \ell_2})$.

The matrix $n^{-1} \Phi_{\ell_1}^\ast \Phi_{\ell_2}$ can thus be computed in $\mathcal{O}(n \log n)$. Consequently, the full matrix $\hat \Sigma$ can be assembled in $\mathcal{O}(d^2 n \log n)$ operations (since there are $d^2$ blocks of $\Phi^\star\Phi$-type).
The linear system involving $(\hat \Sigma + \lambda I)$ can then be solved in $\mathcal{O}(d^3 m^3) = \mathcal{O}(n)$ operations, provided that $s \geq 1$. Overall, this yields a total computational complexity of $\mathcal{O}(d^2 n \log n)$, while remaining fully compatible with GPU acceleration. This is significantly more efficient than GAMs, which require $\mathcal{O}(d^2 n m^2)$ operations and cannot benefit from NUFFT acceleration due to their reliance on spline bases rather than Fourier representations \citep{wood2014generalized}. In addition, the memory footprint of our Fourier-based kernel method is $\mathcal{O}(n \log n + m^2 d^2) = \mathcal{O}(n \log n)$, while GAMs require $\mathcal{O}(n d m)$ memory to store the design matrix, making them more memory-intensive at scale.

\paragraph{Experiment.} 
In this example, we consider the following additive regression setting in dimension $d = 5$:
\begin{itemize}
    \item[(i)] The input variable $X$ is uniformly distributed over the hypercube $\Omega = ]0, 1[^5$;
    \item[(ii)]The response is given by
    \[
    Y = \sum_{\ell=1}^5 \Big( \exp\big( \frac{X_\ell}{\ell + 1} \big) - 1 \Big) + \varepsilon,
    \]
    where $\varepsilon \sim \mathcal{N}(0,1)$ is independent of $X$.
\end{itemize}
We implement the low-bias additive kernel regression estimator with smoothness parameter $s = 2$, and evaluate its performance over $16$ different values of $n$ ranging from $1$ to $10^8$.

\begin{figure}
    \centering
    \includegraphics[width=.45\linewidth]{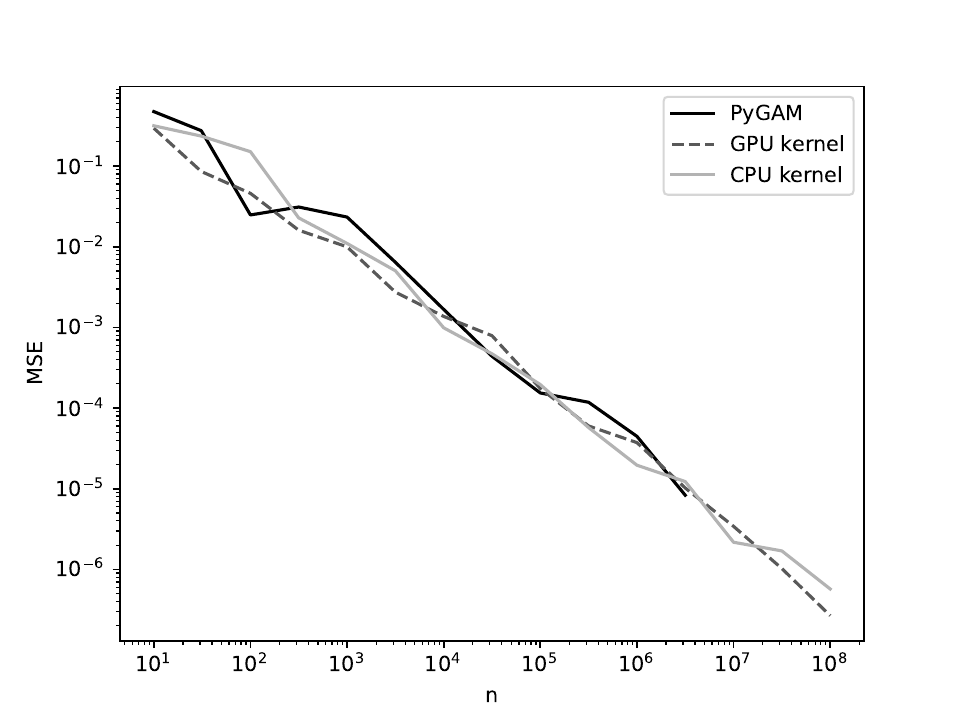}
    \includegraphics[width=.45\linewidth]{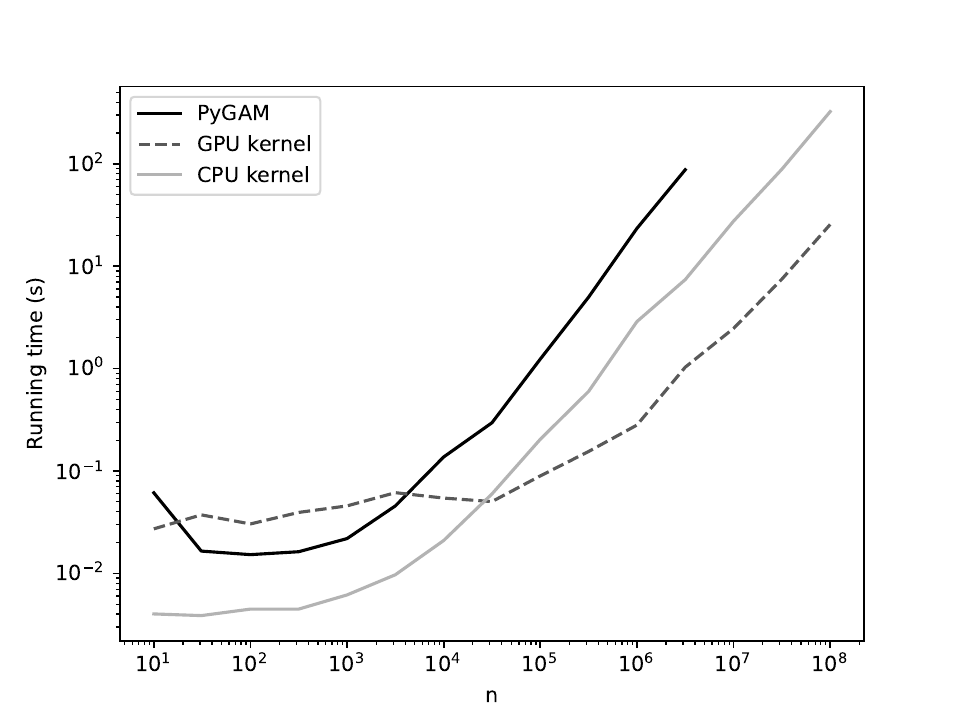}
    \caption{Additive model with $d=5$}
    \label{fig:add}
\end{figure}

In Figure~\ref{fig:add}, we compare our low-bias additive kernel estimator implemented on CPU (CPU kernel) and on a NVIDIA T4 GPU (GPU kernel) with the \texttt{PyGAM} package, which runs on CPU only \citep{serven2018pygam}. For a fair comparison, the \texttt{PyGAM} estimator is given $2m + 1$ splines for each of the five univariate components, where $m = 1 + \lfloor n^{-1/(2s + 1)} / d \rfloor$ with $s = 2$. This ensures that both the kernel and spline-based estimators use the same number of parameters. In both methods, the regularization parameter is set to $\lambda_n = n^{-2s/(2s + 1)}$.

Figure~\ref{fig:add} (Left) shows that both estimators achieve comparable predictive performance, with our kernel estimators performing slightly better than \texttt{PyGAM}. Figure~\ref{fig:add} (Right) highlights the computational advantages of our method. The kernel estimators exhibit significantly faster runtimes compared to \texttt{PyGAM}, especially as $n$ increases. In particular, \texttt{PyGAM} exceeds the available CPU memory (15 GB) at $n = 10^7$, making it infeasible to evaluate for larger datasets. In contrast, our GPU implementation remains scalable and efficient. For instance, at $n = 10^8$, the GPU kernel estimator is approximately 14 times faster than its CPU counterpart.

\begin{figure}
    \centering
    \includegraphics[width=.45\linewidth]{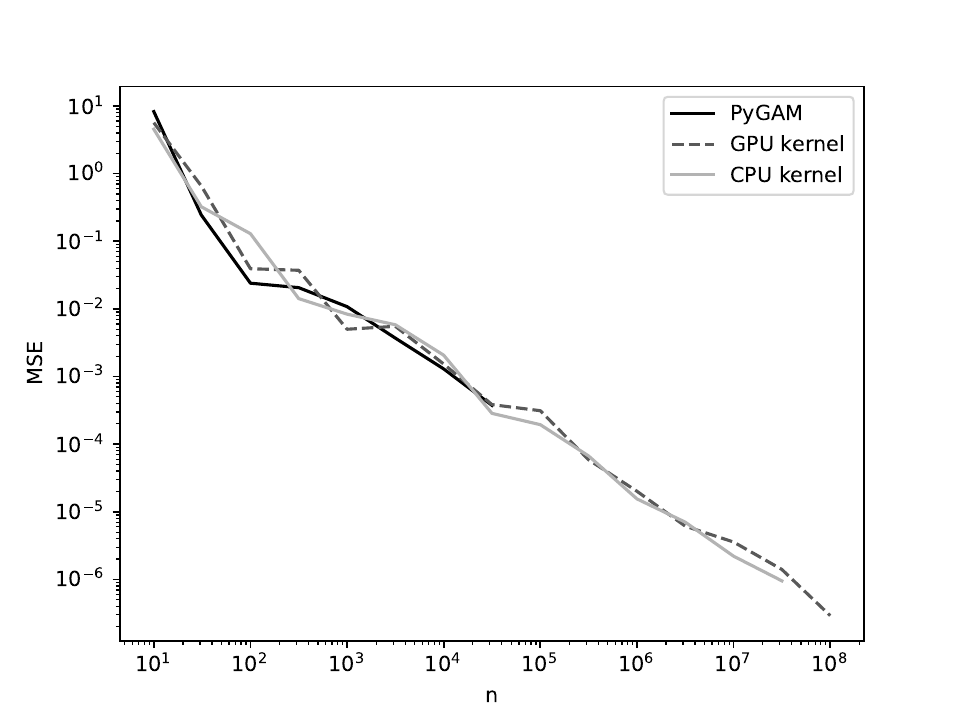}
    \includegraphics[width=.45\linewidth]{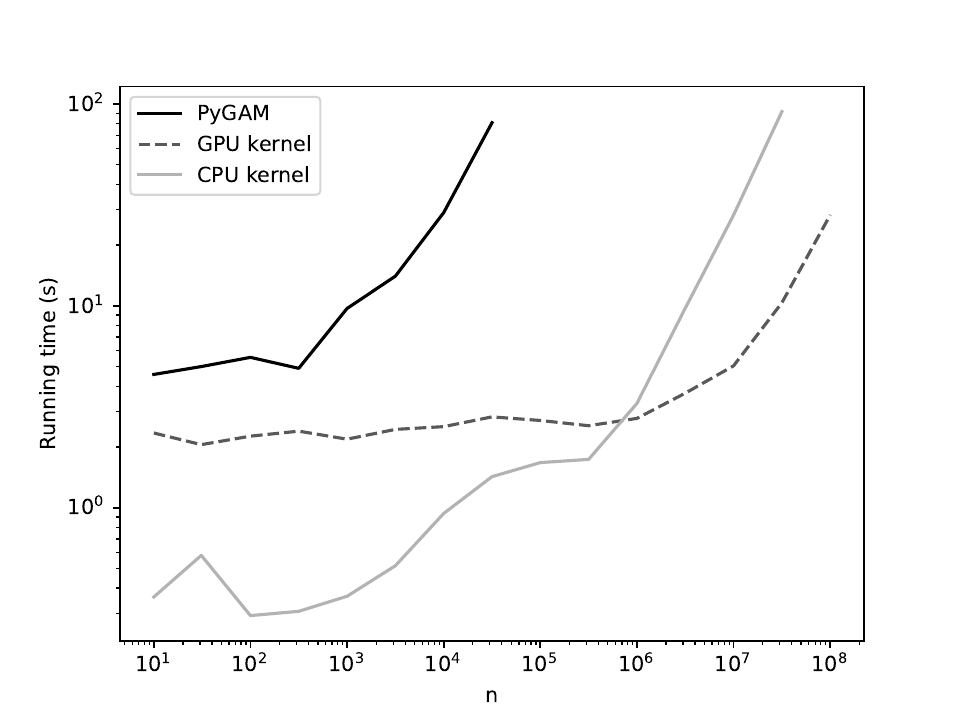}
    \caption{Grid search with $300$ hyperparameters for the additive model with $d=5$}
    \label{fig:add-gs}
\end{figure}

\paragraph{Grid search for $\lambda$.}
Interestingly, the most computationally demanding step in the kernel estimation pipeline is the NUFFT computation, which has a complexity of $\mathcal{O}(d^2 n \log n)$. In contrast, the subsequent matrix inversion step is significantly faster, with a complexity of $\mathcal{O}(d^3 m^3) = \mathcal{O}(d^3 n^{-3/5})$ when $s = 2$. This observation is particularly relevant because it implies that multiple values of the regularization parameter $\lambda$ can be tested efficiently, making grid search highly tractable. Figure~\ref{fig:add-gs} compares the \texttt{gam.gridsearch} function from \texttt{PyGAM} to our kernel method implemented on both CPU and GPU, using a grid of 300 candidate values for $\lambda$. While the runtime of the \texttt{PyGAM} grid search becomes prohibitive at $n = 10^5$, our GPU-based kernel implementation completes the entire grid search in under 30 seconds even for $n = 10^8$.

\section*{Conclusion}
In this work, we have introduced a scalable framework for kernel methods with $\mathcal{O}(n \log n)$ complexity, designed to fully exploit GPU acceleration. The approach combines a Fourier representation of the kernel with the non-uniform fast Fourier transform (NUFFT), yielding fast and memory-efficient computations. We instantiate the framework for Sobolev, physics-informed, and additive kernel models, thereby demonstrating its flexibility and broad applicability. From a theoretical perspective, we establish that the proposed Sobolev and additive kernels achieve optimal minimax convergence rates, in line with classical results from kernel theory.

\bibliography{biblio}

\begin{thebibliography}{17}
\providecommand{\natexlab}[1]{#1}
\providecommand{\url}[1]{\texttt{#1}}
\expandafter\ifx\csname urlstyle\endcsname\relax
  \providecommand{\doi}[1]{doi: #1}\else
  \providecommand{\doi}{doi: \begingroup \urlstyle{rm}\Url}\fi

\bibitem[Agranovich(2015)]{agranovich2015lispchitz}
Mikhail~S. Agranovich.
\newblock \emph{Sobolev Spaces, Their Generalizations and Elliptic Problems in
  Smooth and Lipschitz Domains}.
\newblock Springer, Cham, 2015.

\bibitem[Bach(2024)]{bach2024learning}
Francis Bach.
\newblock \emph{Learning Theory from First Principles}.
\newblock MIT Press, 2024.

\bibitem[Doum{\`e}che et~al.(2024)Doum{\`e}che, Bach, Biau, and
  Boyer]{doumeche2024physicsinformed}
Nathan Doum{\`e}che, Francis Bach, G{\'e}rard Biau, and Claire Boyer.
\newblock Physics-informed machine learning as a kernel method.
\newblock In Shipra Agrawal and Aaron Roth (eds.), \emph{Proceedings of Thirty
  Seventh Conference on Learning Theory}, volume 247 of \emph{Proceedings of
  Machine Learning Research}, pp.\  1399--1450. PMLR, 2024.

\bibitem[Doumèche et~al.(2024)Doumèche, Bach, Biau, and
  Boyer]{doumèche2024physicsinformedkernellearning}
Nathan Doumèche, Francis Bach, Gérard Biau, and Claire Boyer.
\newblock Physics-informed kernel learning, 2024.
\newblock URL \url{https://arxiv.org/abs/2409.13786}.

\bibitem[Doumèche et~al.(2025)Doumèche, Bach, Éloi Bedek, Biau, Boyer, and
  Goude]{doumèche2025forecastingtimeseriesconstraints}
Nathan Doumèche, Francis Bach, Éloi Bedek, Gérard Biau, Claire Boyer, and
  Yannig Goude.
\newblock Forecasting time series with constraints, 2025.
\newblock URL \url{https://arxiv.org/abs/2502.10485}.

\bibitem[Golub \& Loan(2013)Golub and Loan]{golub2013matrix}
Gene~Howard Golub and Charles F.~Van Loan.
\newblock \emph{Matrix computations}.
\newblock The Johns Hopkins University Press, 2013.

\bibitem[Hastie \& Tibshirani(1986)Hastie and
  Tibshirani]{hastie1986generalized}
Trevor Hastie and Robert Tibshirani.
\newblock {Generalized additive models}.
\newblock \emph{Statistical Science}, 1:\penalty0 297--310, 1986.

\bibitem[Karniadakis et~al.(2021)Karniadakis, Kevrekidis, Lu, Wang, Perdikaris,
  and Yang]{karniadakis2021piml}
George~Em Karniadakis, Ioannis~G. Kevrekidis, Lu~Lu, Sifan Wang, Paris
  Perdikaris, and Liu Yang.
\newblock Physics-informed machine learning.
\newblock \emph{Nature Reviews Physics}, 3:\penalty0 422--440, 2021.

\bibitem[Lee(1986)]{lee1986fast}
David Lee.
\newblock Fast multiplication of a recursive block toeplitz matrix by a vector
  and its application.
\newblock \emph{Journal of Complexity}, 2\penalty0 (4):\penalty0 295--305,
  1986.

\bibitem[Meanti et~al.(2020)Meanti, Carratino, Rosasco, and
  Rudi]{Meanti2020kernel}
Giacomo Meanti, Luigi Carratino, Lorenzo Rosasco, and Alessandro Rudi.
\newblock Kernel methods through the roof: Handling billions of points
  efficiently.
\newblock In H.~Larochelle, M.~Ranzato, R.~Hadsell, M.F. Balcan, and H.~Lin
  (eds.), \emph{Advances in Neural Information Processing Systems}, volume~33,
  pp.\  14410--14422. Curran Associates, Inc., 2020.

\bibitem[Mourtada \& Rosasco(2022)Mourtada and Rosasco]{mourtada2022an}
Jaouad Mourtada and Lorenzo Rosasco.
\newblock An elementary analysis of ridge regression with random design.
\newblock \emph{Comptes Rendus. Mathématique}, 360:\penalty0 1055--1063, 2022.

\bibitem[Nemirovski(2000)]{Nemirovski2000estimating}
Arkadi Nemirovski.
\newblock \emph{Estimating signals satisfying differential inequalities}, pp.\
  155--182.
\newblock Springer Berlin Heidelberg, Berlin, Heidelberg, 2000.

\bibitem[Servén \& Brummitt(2018)Servén and Brummitt]{serven2018pygam}
Daniel Servén and Charlie Brummitt.
\newblock pygam: Generalized additive models in python, March 2018.
\newblock URL \url{https://doi.org/10.5281/zenodo.1208723}.

\bibitem[Shih et~al.(2021)Shih, Wright, Anden, Blaschke, and
  Barnett]{shih2021nonuniform}
Yu-hsuan Shih, Garrett Wright, Joakim Anden, Johannes Blaschke, and Alex~H.
  Barnett.
\newblock { cuFINUFFT: a load-balanced GPU library for general-purpose
  nonuniform FFTs }.
\newblock In \emph{2021 IEEE International Parallel and Distributed Processing
  Symposium Workshops (IPDPSW)}, pp.\  688--697, Los Alamitos, CA, USA, 2021.
  IEEE Computer Society.

\bibitem[Steinwart \& Christmann(2008)Steinwart and Christmann]{steinwart2008}
Ingo Steinwart and Andreas Christmann.
\newblock \emph{Support Vector Machines}.
\newblock Information Science and Statistics (ISS). Springer New York, NY,
  2008.

\bibitem[Tsybakov(2009)]{tsybakov2009introduction}
Alexandre~B. Tsybakov.
\newblock \emph{Introduction to Nonparametric Estimation}.
\newblock Springer, New York, 2009.

\bibitem[Wood et~al.(2014)Wood, Goude, and Shaw]{wood2014generalized}
Simon~N. Wood, Yannig Goude, and Simon Shaw.
\newblock Generalized additive models for large data sets.
\newblock \emph{Journal of the Royal Statistical Society Series C: Applied
  Statistics}, 64\penalty0 (1):\penalty0 139--155, 2014.

\end{thebibliography}
\bibliographystyle{iclr2025_conference}

\appendix

\section{Proofs}
\subsection{Proof of Proposition~\ref{lem:fourier_approx}}
For $f\in H^s(\Omega)$, let $\theta(f) \in \mathbb C^{Z^d}$ be the Fourier coefficients of $f$, as in Proposition~\ref{prop:fourier_rep}.
To simplify notation, we write $\theta = \theta(f)$.
Let $f_{m}$ be the projection of $f$ on $H_{m}$, defined by $$f_{m}(x) = \sum_{\|k\|_\infty \leq m} \theta_k \exp(i\pi \langle k, x\rangle/2L).$$ 
Parseval's identity states that  
\[\mathbb E(( f(X)-f_{m}(X))^2) \leq \kappa \int_{[-2L,2L]^d}( f(x)-f_{m}(x))^2dx = \kappa (4L)^d \sum_{\|k\|_\infty > m} |\theta_k|^2.\]
Since $f \in H^s(\Omega)$, one has $\sum_{k\in \mathbb Z^d} |\theta_k|^2\|k\|_2^{2s} < \infty$. 
Let $k(\ell)$ be a reindexing of $\mathbb Z^d$  such that $\|k(\ell)\|_2$ is non-decreasing. 
According to \citet[][Proposition~A.7]{doumeche2024physicsinformed}, there is a constant $C_1 >0$ such that $\|k(j)\|_2^{2s} \leq C_1 j^{2s/d}$. Since $\|k\|_\infty > m \Rightarrow \|k\|_2 > m$, we deduce that
\[\sum_{\|k\|_\infty > m} |\theta_k|^2 =  \sum_{\|k\|_\infty > m} (|\theta_k| \|k\|_2^{2s})\|k\|^{-2s}_2 \leq  \sum_{\ell > (2m+1)^d} (|\theta_{k(\ell)}|^2 \|k(\ell)\|_2^{2s})  \|k(\ell)\|_2^{-2s}.\]
Now, consider its Abel transform $A_\ell = \sum_{j = 0}^{\ell-1} (|\theta_{k(j)}|^2 \|k(j)\|_2^{2s}) $.
By construction, $A_\ell$ is bounded by $C_\Omega\|f\|_{H^s(\Omega)}^2$.
Note that 
\[\sum_{\ell > (2m+1)^d} (|\theta_{k(\ell)}|^2 \|k(\ell)\|_2^{2s})  \|k(\ell)\|_2^{-2s} = \sum_{\ell > (2m+1)^d} (A_{\ell+1}-A_\ell)  \|k(\ell)\|_2^{-2s}.\]
This telescopic series satisfies the Abel property (discrete version of the integration by parts), i.e., 
\[ \sum_{\ell > (2m+1)^d} (A_{\ell+1}-A_\ell)  \|k(\ell)\|_2^{-2s} = A_\ell \|k((2m+1)^d)\|_2^{-2s}+\sum_{\ell > (2m+1)^d} A_{\ell+1} ( \|k(\ell)\|_2^{-2s}- \|k(\ell+1)\|_2^{-2s}).\]
Moreover, since $\|k(\ell)\|_2$ is non-decreasing, the right-hand term is bounded by
\begin{align*}
    \sum_{\ell > (2m+1)^d} A_{\ell+1} ( \|k(\ell)\|_2^{-2s}- \|k(\ell+1)\|_2^{-2s}) &\leq C_\Omega \|f\|_{H^s(\Omega)}^2 \sum_{\ell > (2m+1)^d}  \|k(\ell)\|_2^{-2s}- \|k(\ell+1)\|_2^{-2s} \\
    &= C_\Omega \|f\|_{H^s(\Omega)}^2 \|k((2m+1)^{d})\|_2^{-2s}.
\end{align*}
Overall, we deduce that 
\begin{align*}
    \sum_{\|k\|_\infty > m} |\theta_k|^2 &\leq 2 C_\Omega \|f\|_{H^s(\Omega)}^2 \|k((2m+1)^{d})\|_2^{-2s}\\
    &\leq C_1 C_\Omega \|f\|_{H^s(\Omega)}^2 (2m+1)^{-2sd/d}\\ 
    &\leq C \|f\|_{H^s(\Omega)}^2 m^{-2s}, 
\end{align*}
where $C = C_1 C_\Omega (4L)^d$.
\subsection{Proof of Propositions~\ref{prop:sob_ini} and \ref{prop:sob_rest}}

\paragraph{General kernel regression setting.}
Let $\theta\in \mathbb C^{(2m+1)^d}$ be an arbitrary vector of Fourier coefficients.
 We are interested in bounding the quantity
\begin{align}
    \mathbb E(\|f_{\hat \theta}-f^\star\|_{L^2(\mathbb P_X)}^2) &= \mathbb E(\|(f_{\hat \theta}-f_{\theta}) + (f_{\theta}-f^\star)\|_{L^2(\mathbb P_X)}^2)\nonumber \\
    &\leq 2\mathbb E(\|f_{\hat \theta}-f_{\theta}\|_{L^2(\mathbb P_X)}^2) + 2\mathbb E(\|f_{\theta}-f^\star\|_{L^2(\mathbb P_X)}^2).\label{eq:initial}
\end{align} 
Let us start by bonding 
the error term $\mathbb E(\|f_{\hat \theta}-f_{\theta}\|_{L^2(\mathbb P_X)}^2)$.

\paragraph{Generalization error.} In both propositions, $\hat \theta$ can be written as 
$$\hat \theta = (\hat \Sigma + \lambda R)^{-1} \Phi^\star \mathbb Y/n,$$
where  $\hat \Sigma = n^{-1}\Phi^\star \Phi$ is the empirical covariance matrix and $R$ is a regularization matrix. 
($R = S^2$ in the Sobolev regression, and  $R = I$ for the low-bias Sobolev regression.)
Let $\Sigma = \mathbb E(\hat \Sigma)$ be the theoretical covariance matrix, defined by $\langle \theta_1,  \Sigma \theta_2\rangle = \mathbb E(\langle\phi(X), \theta_1\rangle\langle \phi(X), \theta_2\rangle)$. 
Since $\mathbb Y$ can be decomposed as $\mathbb Y = \Phi\theta + \delta \mathbb Y + \epsilon$, where $\epsilon = (\varepsilon_1, \hdots, \varepsilon_n)^\top$ and $\delta \mathbb Y_j = f^\star(X_j)-f_{\theta}(X_j)$, we can write
\begin{align*}
    \hat \theta &= (\hat \Sigma + \lambda R)^{-1} \Phi^\star \mathbb Y/n\\
    &= (\hat \Sigma + \lambda R)^{-1} (\hat \Sigma \theta +\Phi^\star (\delta\mathbb Y+\epsilon)/n)\\
    &= \theta + (\hat \Sigma + \lambda R)^{-1} (- \lambda R \theta +\Phi^\star (\delta\mathbb Y+\epsilon)/n).
\end{align*}
This leads to the following decomposition of the generalization error:
\begin{align*}
    \mathbb E(\|f_{\hat \theta}-f_{\theta}\|_{L^2(\mathbb P_X)}^2) &= \mathbb E(\|\sqrt{\Sigma}(\hat \theta-\theta)\|_2^2)\\
    &= \mathbb E(\|\sqrt{\Sigma}(\hat \Sigma + \lambda R)^{-1}(-\lambda R\theta+\Phi^*(\epsilon+\delta \mathbb Y)/n )\|_2^2)\\
    &\leq 3\mathbb E(\|\sqrt{\Sigma}(\hat \Sigma + \lambda R)^{-1}\lambda R\theta\|_2^2)   + 3\mathbb E(\|\sqrt{\Sigma}(\hat \Sigma + \lambda R)^{-1}\Phi^*\epsilon/n \|_2^2)\\
    &\quad + 3\mathbb E(\|\sqrt{\Sigma}(\hat \Sigma + \lambda R)^{-1}\Phi^*\delta \mathbb Y/n \|_2^2),
\end{align*}
where we used the fact that $(x_1+x_2+x_3)^2\leq 3x_1^2+3x_2^2+3x_3^2$. 
The first term $\mathbb E(\|\sqrt{\Sigma}(\hat \Sigma + \lambda R)^{-1}\lambda R\theta\|_2^2)$ is a bias term depending on the regularization $\lambda R$. 
The second term $\mathbb E(\|\sqrt{\Sigma}(\hat \Sigma + \lambda R)^{-1}\Phi^*\epsilon/n \|_2^2)$ is a variance term depending on the noise $\epsilon$. 
The last term $\mathbb E(\|\sqrt{\Sigma}(\hat \Sigma + \lambda R)^{-1}\Phi^*\delta \mathbb Y/n \|_2^2)$ is another approximation error term, which measures the impact of the interaction between the regularization $R$ and the approximation errors $\delta \mathbb Y$. 
In what follows, we will successively bound these three errors terms.

\paragraph{Bias term.} 
According to  \citet[Lemma~7.1,][]{bach2024learning}, one has
\begin{align*}
    \mathbb E(\|\sqrt{\Sigma}(\hat \Sigma + \lambda R)^{-1}\lambda R\theta\|_2^2) &= \lambda^2 \mathbb E(\|\sqrt{\Sigma}R^{-1/2}(R^{-1/2}\hat \Sigma R^{-1/2} + \lambda )^{-1} R^{1/2}\theta\|_2^2)\\
    &\leq \lambda (1+\frac{\alpha^2}{\lambda n})^2 \langle R^{1/2}\theta, R^{-1/2} \Sigma R^{-1/2}(R^{-1/2} \Sigma R^{-1/2} + \lambda )^{-1} R^{1/2}\theta\rangle\\
    &\leq \lambda (1+\frac{\alpha^2}{\lambda n})^2 \|R^{1/2}\theta\|_2^2,
\end{align*} 
where $\alpha = \max_{x\in \mathbb R^d} \|R^{-1/2}\phi(x)\|_2$. Observe that $R^{1/2} = S$ for the Sobolev regression and $R^{1/2} = I$ for the low-bias Sobolev regression. 
Thus, $\alpha^2 = \sum_{k\in \mathbb Z^d} \frac{1}{1+\|k\|^{2s}_2} < \infty$ for the Sobolev regression, while $\alpha^2 = (2m+1)^d$ for the low-bias Sobolev regression. 
Remark that $\|R^{1/2}\theta\|_2^2$ is the kernel norm.  

\paragraph{Variance term.}
By expanding $\epsilon = \sum_{j=1}^n \varepsilon_j e_j$ on the canonical basis $(e_1, \hdots, e_n)$, we have that
\begin{align*}
    &\mathbb E(\|\sqrt{\Sigma}(\hat \Sigma + \lambda R)^{-1}\Phi^*\epsilon/n) \|_2^2) \\
    &= n^{-2}\sum_{j_1, j_2 = 1}^n   \mathbb E( e_{j_1}^\star \Phi (\hat \Sigma + \lambda R)^{-1}\Sigma(\hat \Sigma + \lambda R)^{-1}\Phi^*e_{j_2} \mathbb E(\varepsilon_{j_1} \varepsilon_{j_2}\mid X_1, \dots X_n)).
\end{align*}
Since $\varepsilon_{j_1}$ and $\varepsilon_{j_2}$ are independent when $j_1\neq j_2$, and since $\mathbb E(\varepsilon_{j} \mid X_1, \dots X_n)) = 0$, we deduce that 
\begin{align*}
    &\mathbb E(\|\sqrt{\Sigma}(\hat \Sigma + \lambda R)^{-1}\Phi^*\varepsilon/n \|_2^2) 
    \\
    &= \mathbb E(\varepsilon_1^2)n^{-2} \hbox{tr} (\mathbb E( \Phi (\hat \Sigma + \lambda R)^{-1}\Sigma(\hat \Sigma + \lambda R)^{-1}\Phi^*))\\
    &\leq \sigma^2 n^{-2} \hbox{tr} (\mathbb E( \Phi R^{-1/2} (R^{-1/2} \hat \Sigma R^{-1/2} + \lambda )^{-1}R^{-1/2}\Sigma R^{-1/2}(R^{-1/2}\hat \Sigma R^{-1/2} + \lambda )^{-1}R^{-1/2}\Phi^*))\\
    &= \sigma^2 n^{-1} \hbox{tr} (\mathbb E(   (R^{-1/2} \hat \Sigma R^{-1/2} + \lambda )^{-1}R^{-1/2}\Sigma R^{-1/2}(R^{-1/2}\hat \Sigma R^{-1/2} + \lambda )^{-1}R^{-1/2}\hat \Sigma R^{-1/2}))\\
    &\leq  \sigma^2 n^{-1} \hbox{tr} (\mathbb E(   (R^{-1/2} \hat \Sigma R^{-1/2} + \lambda )^{-1}R^{-1/2}\Sigma R^{-1/2})).
\end{align*}
Using \citet[Lemma~7.1,][]{bach2024learning}, we conclude that 
\begin{align*}
    \mathbb E(\|\sqrt{\Sigma}(\hat \Sigma + \lambda R)^{-1}\Phi^*\varepsilon/n \|_2^2) 
    &\leq  \sigma^2 n^{-1} (1+\frac{\alpha^2}{\lambda n}) \hbox{tr} (\mathbb E(   (R^{-1/2} \Sigma R^{-1/2} + \lambda )^{-1}R^{-1/2}\Sigma R^{-1/2}))\\
    &\leq \sigma^2 n^{-1} (1+\frac{\alpha^2}{\lambda n}) (2m+1)^d.
\end{align*}

\paragraph{Approximation term.} 
Let $(X_{n+1}, Y_{n+1})$ be a new sample, drawn  independently from $(X_1, Y_1), \dots, (X_n, Y_n)$.
Let $$C = \sum_{j=1}^{n+1} R^{-1/2}\phi(X_j) \phi(X_j)^\star R^{-1/2},$$ so that $C = nR^{-1/2} \hat \Sigma R^{-1/2} + 
 R^{-1/2} \phi(X_{n+1}) \phi(X_{n+1})^\star R^{-1/2}$. The rationale behind this approach is that $(X_{n+1}, Y_{n+1})$ relates to $\Sigma$ by $R^{-1/2} \Sigma R^{-1/2} = \mathbb E(R^{-1/2} \phi(X_{n+1}) \phi(X_{n+1} )^\star R^{-1/2})$. Moreover,  $\hat \Sigma$ and $C$ are related by  \citet[Equation~(7.26),][]{bach2024learning}, so that 
\begin{align*}
    &(C+\lambda n)^{-1}R^{-1/2} \phi(X_{n+1}) \\
    &\quad = (
1+\langle R^{-1/2} \phi(X_{n+1}), (n R^{-1/2} \hat\Sigma R^{-1/2} +n\lambda)^{-1}R^{-1/2}\phi(X_{n+1})\rangle)^{-1}\\
&\qquad \times (n R^{-1/2} \hat\Sigma R^{-1/2} +n\lambda)^{-1} R^{-1/2} \phi(X_{n+1}),
\end{align*}
where the scaling coefficient is such that 
$$(
1+\langle R^{-1/2} \phi(X_{n+1}), (n R^{-1/2} \hat\Sigma R^{-1/2} +n\lambda)^{-1}R^{-1/2}\phi(X_{n+1})\rangle)^{-1} 
 \geq (1+\frac{\alpha^2}{\lambda n})^{-1} .$$
Thus, 
\begin{align*}
    &\mathbb E(\|\sqrt{\Sigma}(\hat \Sigma + \lambda R)^{-1}\Phi^*\delta \mathbb Y/n \|_2^2)\\
    &\quad = n^{-2} \mathbb E((\delta \mathbb Y) ^\star\Phi(\hat \Sigma + \lambda R)^{-1}\phi(X_{n+1}) \phi(X_{n+1})^\star(\hat \Sigma + \lambda R)^{-1}\Phi^*\delta \mathbb Y )\\
    & \quad= n^{-2} \mathbb E(|\langle \Phi(\hat \Sigma + \lambda R)^{-1}\phi(X_{n+1}), \delta \mathbb Y \rangle|^2)\\
    &\quad= n^{-2} \mathbb E(|\langle \Phi R^{-1/2}( R^{-1/2}\hat \Sigma R^{-1/2}+ \lambda )^{-1} R^{-1/2}\phi(X_{n+1}), \delta \mathbb Y \rangle|^2)\\
    &\quad\leq   (1+\frac{\alpha^2}{\lambda n})^{2} \mathbb E(|\langle \Phi R^{-1/2}(C+\lambda n)^{-1}R^{-1/2} \phi(X_{n+1}), \delta \mathbb Y \rangle|^2)\\
    &\quad\leq   (1+\frac{\alpha^2}{\lambda n})^{2} \mathbb E(\| \Phi R^{-1/2}(C+\lambda n)^{-1}R^{-1/2} \phi(X_{n+1})\|_2^2\;\| \delta \mathbb Y \|^2).
\end{align*}
Since $R^{-1/2} \Phi^\star \Phi R^{-1/2} = n R^{-1/2} \hat \Sigma R^{-1/2} \leq C$, we deduce that 
\begin{align*}
    &\| \Phi R^{-1/2}(C+\lambda n)^{-1}R^{-1/2} \phi(X_{n+1})\|_2^2\\
    &\quad\leq \phi(X_{n+1})^\star R^{-1/2}(C+\lambda n)^{-1}C(C+\lambda n)^{-1}R^{-1/2} \phi(X_{n+1})\\
    &\quad\leq \phi(X_{n+1})^\star R^{-1/2}(C+\lambda n)^{-1}R^{-1/2} \phi(X_{n+1}).
\end{align*}
Thus, 
\begin{align*}
    &\mathbb E(\|\sqrt{\Sigma}(\hat \Sigma + \lambda R)^{-1}\Phi^*\delta \mathbb Y/n \|_2^2)\\
    &\quad\leq (1+\frac{\alpha^2}{\lambda n})^{2} \mathbb E(\|(C+\lambda n)^{-1/2}R^{-1/2} \phi(X_{n+1})\|_2^2\;\| \delta \mathbb Y \|^2)\\
    &\quad\leq (1+\frac{\alpha^2}{\lambda n})^{2} \mathbb E(\|(C+\lambda n)^{-1/2}R^{-1/2} \phi(X_{n+1})\|_2^2\;(\| \delta \mathbb Y \|^2+(\delta Y_{n+1})^2)).
\end{align*}
Since the observations $(X_1, Y_1), \dots, (X_{n+1}, Y_{n+1})$ are assumed i.i.d., the random variables $\|(C+\lambda n)^{-1/2}R^{-1/2} \phi(X_{j})\|_2^2\;(\| \delta \mathbb Y \|^2+(\delta Y_{n+1})^2)$ have the same distribution for all $1\leq j \leq n+1$. Therefore, they also share the same expectation. Thus, 
\begin{align*}
    &\mathbb E(\|(C+\lambda n)^{-1/2}R^{-1/2} \phi(X_{n+1})\|_2^2\;(\| \delta \mathbb Y \|^2+(\delta Y_{n+1})^2))\\&= \frac{1}{n+1}\mathbb E\Big(\sum_{j=1}^{n+1}\|(C+\lambda n)^{-1/2}R^{-1/2} \phi(X_{j})\|_2^2\;(\| \delta \mathbb Y \|^2+(\delta Y_{n+1})^2)\Big)\\&= \frac{1}{n+1}\mathbb E\Big(\sum_{j=1}^{n+1}\hbox{tr}(\phi(X_{j})^\star R^{-1/2}(C+\lambda n)^{-1}R^{-1/2} \phi(X_{j}))\;(\| \delta \mathbb Y \|^2+(\delta Y_{n+1})^2)\Big)\\
    &= \frac{1}{n+1}\mathbb E\Big(\hbox{tr}( (C+\lambda n)^{-1} \sum_{j=1}^{n+1}R^{-1/2}\phi(X_{j})\phi(X_{j})^\star R^{-1/2})\;(\| \delta \mathbb Y \|^2+(\delta Y_{n+1})^2)\Big)\\
    &= \frac{1}{n+1}\mathbb E\Big(\hbox{tr}( (C+\lambda n)^{-1} C)\;(\| \delta \mathbb Y \|^2+(\delta Y_{n+1})^2)\Big)\\
    &\leq \frac{1}{n+1}\mathbb E\Big((\| \delta \mathbb Y \|^2+(\delta Y_{n+1})^2)\Big) = \mathbb E((\delta Y^2)) =\mathbb E(\|f^{\star}-f_{\theta}\|_{L^2(\mathbb P_X)}^2),
\end{align*}
where we use the fact $x = \hbox{tr}(x)$ for any real number $x$, and $\hbox{tr}(M_1 M_2) = \hbox{tr}(M_2 M_1)$.
All in all, we deduce that 
$$\mathbb E(\|\sqrt{\Sigma}(\hat \Sigma + \lambda R)^{-1}\Phi^*\delta \mathbb Y/n \|_2^2)\leq (1+\frac{\alpha^2}{\lambda n})^{2}\; \mathbb E(\|f^{\star}-f_{\theta}\|_{L^2(\mathbb P_X)}^2).$$

\paragraph{Conclusion.} Putting everything together, we deduce from inequality~\ref{eq:initial} that 
\begin{align*}
    \mathbb E(\|f_{\hat \theta}-f^\star\|_{L^2(\mathbb P_X)}^2) &\leq \inf_{\theta \in H_{m}}\Big((2+6(1+\frac{\alpha^2}{\lambda n})^2)\mathbb E(\|f^{\star}-f_{\theta}\|_{L^2(\mathbb P_X)}^2) + \lambda (1+\frac{\alpha^2}{\lambda n})^2 \|R^{1/2}\theta\|_2^2 \Big)\\
    &\qquad + \sigma^2 n^{-1} (1+\frac{\alpha^2}{\lambda n}) (2m+1)^d,
\end{align*}
where 
\begin{itemize}
    \item $\alpha^2 = \sum_{k\in \mathbb Z^d} \frac{1}{1+\|k\|^{2s}_2} < \infty$ for the Sobolev regression, while $\alpha^2 = (2m+1)^d$ for the low-bias Sobolev regression;
    \item  $\|R^{1/2}\theta\|_2^2 = \|S\theta\|_2^2 \leq C_\Omega \|f_\theta\|_{H^s(\Omega)}^2$ for the Sobolev regression, and $\|R^{1/2}\theta\|_2^2 = \|\theta\|_2^2$ for the low-bias Sobolev regression.
\end{itemize}

\paragraph{Convergence rates.} According to Proposition~\ref{prop:fourier_rep}, there is a parameter $\theta(f^\star)$ such that $f^\star(x) = \langle \phi(x), \theta(f^\star)\rangle$. Let $\theta^\star\in \mathbb C^{(2m+1)^d}$ be the truncation of $\theta(f^\star)$, i.e., $\theta^\star_k = \theta(f^\star)_k$.
Proposition~\ref{lem:fourier_approx} states that the approximation error term $\mathbb E(\|f_{\theta^\star}-f^\star\|_{L^2(\mathbb P_X)}^2) = \mathcal{O}(n^{-\gamma})$, where $\gamma = 2s/(2s+d)$. Moreover, the RKHS norm of $\theta^\star$ is bounded by $\max(\|\theta^\star\|_2^2, \|S\theta^\star\|_2^2) \leq \sum_{k\in \mathbb C^{\mathbb Z^d}} (1+\|k\|_2^{2s}) |\theta(f^\star)_k|^2 \leq C_\Omega \|f^\star\|_{H^s(\Omega)}^2$.
Therefore, in the Sobolev regression, setting $\lambda = \mathcal{O}(n^{-2s/(2s+d)})$ such that $\lambda \geq n^{-1}$, and taking $m = n^{1/(2s+d)}$ leads to Sobolev minimax rate. 
Moreover, in the low-bias Sobolev regression, setting $\lambda = \Theta(n^{-2s/(2s+d)})$ and taking $m = n^{1/(2s+d)}$ leads to Sobolev minimax rate. 

\subsection{Proof of Proposition~\ref{prop:additive}}
We rely on the same tools as in the proof of Propositions~\ref{prop:sob_ini} and \ref{prop:sob_rest}.
The key is to remark that the empirical risk 
$$R(\theta) = \|\Phi_1\theta_1+\dots +\Phi_d \theta_d-\mathbb Y\|_2^2 + \lambda \|\theta\|_2^2$$
can be written as 
$$R(\theta) = \|\Phi\theta-\mathbb Y\|_2^2 + \lambda \|\theta\|_2^2$$
where $\Phi = (\Phi_1\mid \dots \mid \Phi_d)$ and $\theta = (\theta_1, \dots, \theta_d)\in \mathbb C^{d(2m+1)}$.
By setting $\varphi(x) = (\phi(x), \dots, \phi(x))\in \mathbb C^{d(2m+1)}$, we have that $f_\theta(x) = \langle \varphi(x), \theta\rangle.$
The empirical covariance matrix is then $\hat \Sigma = \Phi^\star \Phi/n = n^{-1} \sum_{j=1}^n \varphi(X_j) \varphi(X_j)^\star$.

\paragraph{Risk decomposition.} As in the proof of Propositions~\ref{prop:sob_ini} and \ref{prop:sob_rest}, we have, for any $\theta\in \mathbb C^{d(2m+1)}$,  
\begin{align*}
    \mathbb E(\|f_{\hat \theta}-f^\star\|_{L^2(\mathbb P_X)}^2) &\leq 2\mathbb E(\|f_{\hat \theta}-f_{\theta}\|_{L^2(\mathbb P_X)}^2) + 2\mathbb E(\|f_{\theta}-f^\star\|_{L^2(\mathbb P_X)}^2),
\end{align*} 
and 
\begin{align*}
    \mathbb E(\|f_{\hat \theta}-f_{\theta}\|_{L^2(\mathbb P_X)}^2) &\leq 3\mathbb E(\|\sqrt{\Sigma}(\hat \Sigma + \lambda )^{-1}\lambda \theta\|_2^2)   + 3\mathbb E(\|\sqrt{\Sigma}(\hat \Sigma + \lambda )^{-1}\Phi^*\epsilon/n \|_2^2)\\
    &\quad + 3\mathbb E(\|\sqrt{\Sigma}(\hat \Sigma + \lambda )^{-1}\Phi^*\delta \mathbb Y/n \|_2^2),
\end{align*}
where $\Sigma = \mathbb E(\phi(X)\phi(X)^\star)$. 
Similarly, the bias term is bounded by
\begin{align*}
    \mathbb E(\|\sqrt{\Sigma}(\hat \Sigma + \lambda )^{-1}\lambda \theta\|_2^2) \leq \lambda (1+\frac{\alpha^2}{\lambda n})^2 \|\theta\|_2^2,
\end{align*} 
where $\alpha = \max_{x\in \mathbb R^d} \|\phi(x)\|_2 = \sqrt{d(2m+1)}$.
As for the variance term, it is bounded by 
\begin{align*}
    \mathbb E(\|\sqrt{\Sigma}(\hat \Sigma + \lambda )^{-1}\Phi^*\varepsilon/n \|_2^2) 
    &\leq  \sigma^2 n^{-1} (1+\frac{\alpha^2}{\lambda n}) \hbox{tr} (\mathbb E(   ( \Sigma  + \lambda )^{-1}\Sigma)) \leq \sigma^2 \frac{d(2m+1)}{n} (1+\frac{\alpha^2}{\lambda n}).    
\end{align*}
Finally, the approximation term is bounded by 
\begin{align*}
    \mathbb E(\|\sqrt{\Sigma}(\hat \Sigma + \lambda R)^{-1}\Phi^*\delta \mathbb Y/n \|_2^2)\leq (1+\frac{\alpha^2}{\lambda n})^{2}\; \mathbb E(\|f^{\star}-f_{\theta}\|_{L^2(\mathbb P_X)}^2).
\end{align*}
Putting everything together, we obtain
\begin{align*}
    \mathbb E(\|f_{\hat \theta}-f^\star\|_{L^2(\mathbb P_X)}^2) &\leq \inf_{\theta \in \mathbb C^{d(2m+1)}}\Big((2+6(1+\frac{\alpha^2}{\lambda n})^2)\mathbb E(\|f^{\star}-f_{\theta}\|_{L^2(\mathbb P_X)}^2) + \lambda (1+\frac{\alpha^2}{\lambda n})^2 \|\theta\|_2^2 \Big)\\
    &\qquad + \sigma^2 \frac{\alpha^2}{n} (1+\frac{\alpha^2}{\lambda n}).
\end{align*}
\paragraph{Convergence rate.} The convexity of the function $x\mapsto x^2$ leads to
\begin{align*}
    \mathbb E(\|f^{\star}-f_{\theta}\|_{L^2(\mathbb P_X)}^2) &= \mathbb E(\|\sum_{\ell=1}^dg_\ell^{\star}-g_{\theta_\ell}\|_{L^2(\mathbb P_X)}^2) \leq d \sum_{\ell = 1}^d \mathbb E(\|g_\ell^{\star}-g_{\theta_\ell}\|_{L^2(\mathbb P_X)}^2),
\end{align*}
where $g_{\theta_\ell}(x) = \langle \phi_\ell(x), \theta_\ell\rangle$.
According to Proposition~\ref{prop:fourier_rep},  taking $m = n^{1/(2s+1)}$ ensures that 
$$\inf_{\theta_\ell \in \mathbb C^{2m+1}} \mathbb E(\|g_\ell^{\star}-g_{\theta_\ell}\|_{L^2(\mathbb P_X)}^2) + n^{-2s/(2s+1)}\|\theta_\ell\|_2^2 = \mathcal{O}(n^{-2s/(2s+1)}).$$
Thus, choosing $\lambda = n^{-2s/(2s+1)}$ and $m = n^{1/(2s+1)}$, we are led to
$$\inf_{\theta \in \mathbb C^{d(2m+1)}}\Big((2+6(1+\frac{\alpha^2}{\lambda n})^2)\mathbb E(\|f^{\star}-f_{\theta}\|_{L^2(\mathbb P_X)}^2) + \lambda (1+\frac{\alpha^2}{\lambda n})^2 \|\theta\|_2^2 \Big) = \mathcal{O}(n^{-2s/(2s+1)}).$$
We conclude that
$$\mathbb E(\|f_{\hat \theta}-f^\star\|_{L^2(\mathbb P_X)}^2) =  \mathcal{O}(n^{-2s/(2s+1)}).$$
\end{document}